\documentclass{article}

\usepackage{PRIMEarxiv}

\usepackage[utf8]{inputenc} 
\usepackage[T1]{fontenc}    
\usepackage[hidelinks]{hyperref}       
\usepackage{url}            
\usepackage{booktabs}       
\usepackage{amsfonts}       
\usepackage{nicefrac}       
\usepackage{microtype}      
\usepackage{lipsum}
\usepackage{fancyhdr}       
\usepackage{graphicx}

\usepackage{subcaption}
\usepackage{amssymb}
\usepackage{amsmath}
\usepackage{makecell}
\usepackage{threeparttable}

\usepackage{makecell}
\usepackage{tikz}
\usepackage{soul}
\usepackage{adjustbox}
\usepackage{array}
\usepackage{cleveref}
\usepackage{url}
\usepackage{multirow}
\usepackage{float}

\newcolumntype{R}[2]{%
    >{\adjustbox{angle=#1,lap=\width-(#2)}\bgroup}%
    l%
    <{\egroup}%
}

\everypar{\looseness=-1}

\title{Disentanglement by Cyclic Reconstruction}

\author{
  David Bertoin \\
  IRT Saint-Exupéry \\
  ISAE-SUPAERO \\
  ANITI\\
  Toulouse, France\\
  \texttt{david.bertoin@irt-saintexupery.com} \\
   \And
  Emmanuel Rachelson \\
  ISAE-SUPAERO \\
  Université de Toulouse \\
  ANITI\\
  Toulouse, France\\
  \texttt{emmanuel.rachelson@isae-supaero.fr} \\
}

\begin{document}
\maketitle

\begin{abstract}
Deep neural networks have demonstrated their ability to automatically extract meaningful features from data.
However, in supervised learning, information specific to the dataset used for training, but irrelevant to the task at hand, may remain encoded in the extracted representations.
This remaining information introduces a domain-specific bias, weakening the generalization performance.
In this work, we propose splitting the information into a task-related representation and its complementary context representation.
We propose an original method, combining adversarial feature predictors and cyclic reconstruction, to disentangle these two representations in the single-domain supervised case.
We then adapt this method to the unsupervised domain adaptation problem, consisting of training a model capable of performing on both a source and a target domain.
In particular, our method promotes disentanglement in the target domain, despite the absence of training labels. This enables the isolation of task-specific information from both domains and a projection into a common representation. The task-specific representation allows efficient transfer of knowledge acquired from the source domain to the target domain.
In the single-domain case, we demonstrate the quality of our representations on information retrieval tasks and the generalization benefits induced by sharpened task-specific representations.
We then validate the proposed method on several classical domain adaptation benchmarks and illustrate the benefits of disentanglement for domain adaptation. 
\end{abstract}

\section{Introduction}
The wide adoption of Deep Neural Networks in practical supervised learning applications is hindered by their sensitivity to the training data distribution. 
This problem, known as \emph{domain shift}, can drastically weaken, in real-life operating conditions, the performance of a model that seemed perfectly efficient in simulation. 
Learning a model with the goal of making it robust to a specific domain shift is called \emph{domain adaptation} (DA). 
The data available to achieve DA often consist of a labeled training set from a source domain and an unlabeled sample set from a target domain.
This yields the problem of \emph{unsupervised domain adaptation} (UDA).

In this work, we take an information disentanglement perspective on UDA. 
We argue that a key to efficient UDA lies in separating the necessary information to complete the network's task (classification or regression) from a task-orthogonal information which we call context or \emph{style}.
While such a separation appears rather intuitive for samples from the source domain, disentanglement in the target domain seems however a difficult endeavor since the available data is unlabeled. 
Our contribution is two-fold. We propose a formal definition of the disentanglement problem for UDA which, to the best of our knowledge, is new. 
Then we design a new learning method, called DiCyR (Disentanglement by Cyclic Reconstruction), which relies on cyclic reconstruction of inputs in order to achieve efficient disentanglement, including in the target domain. 
We derive DiCyR both in the supervised learning and in the UDA cases.
Although this paper is developed mainly around UDA, we emphasize that our contribution lies in the disentanglement between task and context information. Its consequences carry out to information retrieval in the single domain case as well as to UDA.

This paper is organized as follows. 
Section \ref{sec:problem} presents the required background on supervised learning and UDA, and proposes a definition of disentanglement for UDA. 
Section \ref{sec:related} reviews recent works in the literature that allow for a critical look at our contribution and put it in perspective. 
Section \ref{sec:contrib} introduces DiCyR, first for the single-domain case, and then for the UDA problem. 
Section \ref{sec:xp} empirically evaluates DiCyR against state-of-the-art methods and discusses its strengths, weaknesses, and variants. 
Section \ref{sec:conclusion} summarizes and concludes this paper.

\section{A definition of information disentanglement}
\label{sec:problem}

In this section, we introduce the notations and background upon which we build the contributions of Section \ref{sec:contrib}. 
Let $\mathcal{X}$ be an input space of descriptors and $\mathcal{Y}$ an output space of labels.
A supervised learning problem is defined by a distribution $p_s(x,y)$ over elements of $\mathcal{X}\times\mathcal{Y}$. 
In what follows, $p_s$ will be called the \emph{source} distribution.
One wishes to estimate a mapping $\hat{f}$ that minimizes a risk function of the form $\mathbb{E}_{(x,y)\sim p_s}[l(\hat{f}(x),y)]$, where $l(\hat{y},y)$ is a loss function. 
The optimal estimator is denoted $f$ and one often writes the distribution $\mathbb{P}(y|x)$ as $y\sim f(x)+\eta$, where $\eta$ captures the deviations between $y$ and $f(x)$.
Hence, one tries to learn $f$.
In practice, the risk can only be approximated using a finite set of samples $\{(x_i,y_i)\}_{i=1}^n$ all independently drawn from $p_s$ and $\hat{f}$ is a parametric function (such as a deep neural network) of the form $y=\hat{f}(x;\theta)$.

Domain adaptation (DA) consists in considering a \emph{target} distribution $p_t$ over $\mathcal{X}\times\mathcal{Y}$ that differs from $p_s$, and the transfer of knowledge from learning in the source domain ($p_s$) to the target domain ($p_t$). 
Specifically, \emph{unsupervised} DA exploits the knowledge of a labelled training set $\{(x^{s}_i, y^{s}_i)\}^{n}_{i=1}$ sampled according to $p_s$, and an unlabelled data set $\{(x^{t}_i)\}^{m}_{i=1}$ sampled according to $p_t$.
For instance, the source domain data could be a set of labelled photographs of faces, and the target domain data, a set of unlabelled face photographs, taken with a different camera under different exposure conditions.
The problem consists in minimizing the target loss $\mathbb{E}_{(x,y)\sim p_t}[l(\hat{f}(x),y)]$.

We suppose that a necessary condition to benefit from the knowledge available in the source domain and transfer it to the target domain is the existence of a common information manifold between domains, where an input's projection is sufficient to predict the labels. 
We refer to this intuitive condition as the underlying UDA hypothesis, which differs from the more general case of representation learning \cite{bengio2013representation}.
We call the useful information \emph{task-specific} or \emph{task-related}. 
The complementary information should be called \emph{task-orthogonal}; it is composed of information present in the input but irrelevant to the task at hand. For the sake of naming simplicity, we will call this information \emph{style} (note that it may be empty). 

\looseness=-1
Let $\Pi_\tau: \mathcal{X} \rightarrow \mathcal{T}$ and $\Pi_\sigma: \mathcal{X} \rightarrow \mathcal{S}$ denote two projection operators, where $\mathcal{T}$ and $\mathcal{S}$ denote respectively the latent task-related information space and the latent style-related information space. 
Given the random variable $X$ defined over $\mathcal{X}$, let $T$ and $S$ be the corresponding random variables defined over $\mathcal{T}$ and $\mathcal{S}$. 
Let $\Pi$ be the joint projection $\Pi(x) = \left( \Pi_\tau(x), \Pi_\sigma(x) \right)$.
Conversely, we shall note $\bar{\Pi}:\mathcal{T} \times \mathcal{S} \rightarrow \mathcal{X}$ a reconstruction operator. 
And finally, $c: \mathcal{T} \rightarrow \mathcal{Y}$ will denote the labeling operator which only uses information from $\mathcal{T}$. 
Let also $I(A,B|C)$ denote the mutual information between random variables $A$ and $B$, conditioned by $C$.
We consider that the information of the elements of $\mathcal{X}$ is correctly disentangled by $\Pi=(\Pi_\tau,\Pi_\sigma)$ if one can find $\bar{\Pi}$ and $c$ such that:

\begin{enumerate}
    \item[C1:] $c,\Pi_\tau$ maximize $I(T,Y)$,
    \item[C2:] $\Pi_\tau,\Pi_\sigma$ maximize $I((T,S),X)$,
    \item[C3:] $I(T,S|X)=0$
    \item[C4:] $\Pi_\sigma$ maximizes $I(S,X)$
\end{enumerate}

Condition C1 imposes that the projection into $\mathcal{T}$ retains enough information to correctly label samples.
Condition C2 imposes that all the information necessary for the reconstruction is preserved by the separation performed by $\Pi$.
Condition C3 states that no information is present in both $\mathcal{T}$ and $\mathcal{S}$. 
It can be reformulated as ``$\Pi_\tau,\Pi_\sigma$ minimize $I(T,S|X)$''. 
Conditions C1 to C3 tolerate representations $(\mathcal{T},\mathcal{S})$ that push more than the necessary task information into $\mathcal{T}$. 
By maximizing the mutual information between the style and the image, and since condition C3 imposes no shared information between the task and style representations, condition C4 guarantees that \emph{only} the strictly necessary task information is in $\mathcal{T}$ and all the rest is in $\mathcal{S}$. 
Overall, this formulation of disentanglement boils down to a multi-objective optimization problem on the quadruplet $\langle \Pi_\tau, \Pi_\sigma, \bar{\Pi},c \rangle$.
The existence of a common information manifold across domains that permits seamless generalization from one domain to the other supposes the existence of at least one solution that is dominant for all optimization criteria independently. 
Hence, under the underlying UDA hypothesis, that there exist a classifier that solves the decision problem at hand whatever the considered domain, this multi-objective problem degenerates to a single objective one.
Note that this definition is not restricted to the problem of DA and proposes a formulation for information disentanglement in the general case.
To the best of our knowledge, this is the first explicit formulation of disentanglement as an optimization problem.

\section{Related work}
\label{sec:related}
Disentanglement between the domain-invariant, task-related information   and the domain-specific, task-orthogonal,  style information is a desirable property to have for DA. In the next paragraphs, we cover important work in representation disentanglement, domain adaptation, and their interplay. 
For each contribution, we evaluate whether it complies with conditions C1 to C4, and how.

Before deep learning became prevalent, \cite{tenenbaum2000separating} presented a method using bi-linear models able to separate style from content. More recently, methods based on generative models have demonstrated the ability to disentangle factors of variations from elements of a single domain \cite{rifai2012disentangling,mathieu2016disentangling,chen2016infogan,higgins2017beta,sanchez2019learning}. In a cross-domain setting, \cite{gonzalez2018image} use pairs of images with the same labels from different domains to separate representations into a shared information common to both domains and a domain-exclusive information.
We note that these approaches do not explicitly aim at respecting all conditions listed in Section \ref{sec:problem}. 
Additionally, most require labeled datasets (and in some cases even paired datasets) and thus do not address the \emph{unsupervised} DA problem.

One approach to UDA consists in aligning the source and target distributions statistics, a topic closely related to \emph{batch normalization} \cite{ioffe2015batch}. 
CORAL \cite{sun2017correlation} minimizes the distance between the covariance matrices of the features extracted from the source and target domains. 
Assuming the domain-specific information is contained inside the batch normalization layers, AdaBN \cite{li2016revisiting} aligns the batch statistics by adopting a specific normalization for each domain. 
Autodial \cite{cariucci2017autodial} aims to align source and target feature distributions to a reference one and introduce domain alignment layers to automatically learn the degree of feature alignment needed at different levels of the network. 
Similarly, DWT \cite{roy2019unsupervised} replaces batch normalization layers with domain alignment layers implementing a so-called feature whitening.  
A significant asset of these methods is the possibility to be used jointly with other DA methods (including the one we propose in Section \ref{sec:contrib}).
These methods jointly learn a common representation for elements from both domains.
Conversely, SHOT \cite{liang2020we} freezes the representations learned in the source domain before training a target-specific encoder to align the representations of the target elements by maximizing the mutual information between intermediate feature representations and outputs of the classifier.

Ensemble methods have also been applied to UDA \cite{laine2017temporal,tarvainen2017mean}.
SEDA \cite{french2018selfensembling} combines stochastic data augmentation with self-ensembling to minimize the prediction differences between a student and a teacher network in the target domain. 

Another approach involves learning domain-invariant features, that do not allow to discriminate whether a sample belongs to the source or target domain, while still permitting accurate labeling in the source domain.
This approach relies on the assumption that such features allow efficient labeling in the target domain.
DRCN \cite{ghifary2016deep} builds a two-headed network sharing common layers; one head performs classification in the source domain, while the second is a decoder that performs reconstruction for target domain elements.
\cite{ganin2016domain} propose the DANN method and introduce Gradient Reversal Layers to connect a domain discriminator and a feature extractor. These layers invert the gradient sign during back-propagation so that the feature extractor is trained to fool the domain discriminator.
WDGRL \cite{shen2018wasserstein} modifies DANN and replaces the domain discriminator with a network that approximates the Wasserstein distance between domains.
ADDA \cite{tzeng2017adversarial} optimizes, in an adversarial setting, a generator and a discriminator with an inverted label loss. 

Other methods focus on explicitly disentangling an information shared between domains (analogous to the domain-invariant features above) from a domain-specific information.
Inspired by InfoGAN \cite{chen2016infogan}, CDRD \cite{liu2018detach} isolate a latent factor, representing the domain information, from the rest of an encoding, by maximizing the mutual information between generated images and this latent factor.
Some domain information may still be present in the remaining part of the encoding  and thus may not comply with conditions C3 and C4.

UFDN \cite{liu2018unified} (and also \cite{li2020unsupervised}) trains an encoder to produce domain-invariant representations used by an image generator trained to fool a discriminator with cross-domain images.
DSN \cite{bousmalis2016domain} also produces domain-invariant features by training a shared encoder to fool a domain discriminator.
It trains two domain-private encoders with a difference loss that encourages orthogonality between the shared and the private representations (similarly to condition C3).
DIDA \cite{cao2018dida} (but also \cite{cai2019learning} and \cite{peng2019domain}) combines a domain discriminator with an adversarial classifier to separate the information shared between domains from the domain-specific information.

All these methods build a shared representation that prevents discriminating between source and target domains, while retaining enough information to correctly label samples from the source domain. 
However, because they rely on an adversarial classifier that requires labeled data, they do not guarantee that the complementary, domain-specific information for samples \emph{in the target domain} does not contain information that overlaps with the shared representation. 
In other words, they only enforce C3 in the source domain.
They rely on the assumption that the disentanglement will still hold when applied to target domain elements, which might not be true.

Another identified weakness in methods that achieve a domain-invariant feature space is that their representations might not allow for accurate labeling in the target domain. Indeed, feature alignment does not necessarily imply a correct mapping between domains.
To illustrate this point, consider a binary classification problem (classes $c_1$ and $c_2$) and two domains ($d_1$ and $d_2$). 
Let $(c_1,d_1)$ denote samples of class $c_1$ in $d_1$. 
It is possible to construct an encoding that projects $(c_1,d_1)$ and $(c_2,d_2)$ to the same feature values. 
The same holds for $(c_1,d_2)$ and $(c_2,d_1)$ for different feature values.
This encoding allows discriminating between classes in $d_1$. 
It also fools a domain discriminator since it does not allow predicting the original domain of a projected element. 
However, applying the classification function learned on $d_1$ to the projected $d_2$ elements leads to catastrophic predictions.

Transforming a sample from one domain to the other, while retaining its label information can be accomplished by image-to-image translation methods.
Using an adversarial setting, SBADA-GAN  \cite{russo2018source}, ACAL \cite{hosseini-asl2018augmented}, and CyCADA \cite{hoffman2018cycada} extend the cycle consistency introduced in CycleGAN \cite{zhu2017unpaired}.
A major drawback of these methods lies in the possible instability during training that is caused by the min-max optimization problem induced by the adversarial training of generators and discriminators.

In the next section, we introduce a method that does not rely on a domain discriminator and an adversarial label predictor, but directly minimizes the information sharing between representations.
This allows to guarantee that there is no information redundancy between the task-related and the task-orthogonal style information in both the source and the target domains. 
Along the way, it provides an efficient mechanism to disentangle the task-related information from the style information in the single domain case.   
Our method combines information disentanglement, intra-domain and cross-domain cyclic consistency, to enforce a more principled mapping between each domain.
\section{Disentanglement with Gradient Reversal Layers and cyclic reconstruction}
\label{sec:contrib}
First, we propose an original method to disentangle the task-related information from the style information  for a single domain in a supervised learning setting. In a second step, we propose an adaptation of this method to learn these disentangled representations in both domains for UDA. This disentanglement allows, in turn, to efficiently predict labels in the target domain.

\subsection{Task-style disentanglement in a single domain}
\begin{figure}[h]
\centering
  \begin{subfigure}{0.4\textwidth}
  \begin{center}
    \includegraphics[height=4.5cm]{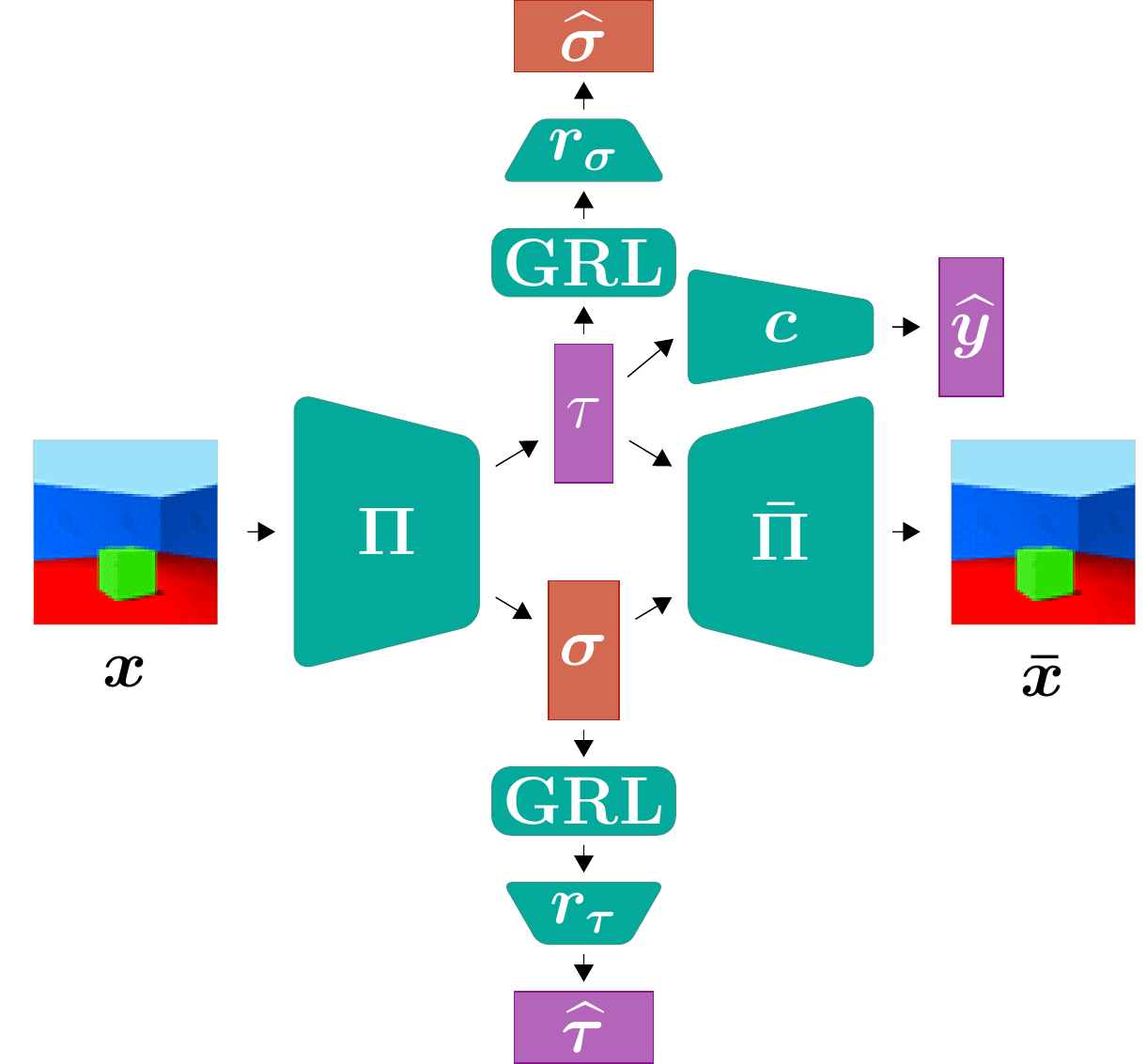}
    \end{center}
    \caption{Supervised learning}
    \label{fig:archi_supervised}
  \end{subfigure}
  \begin{subfigure}{0.45\textwidth}
    
    \begin{center}
        \includegraphics[height=4.5cm]{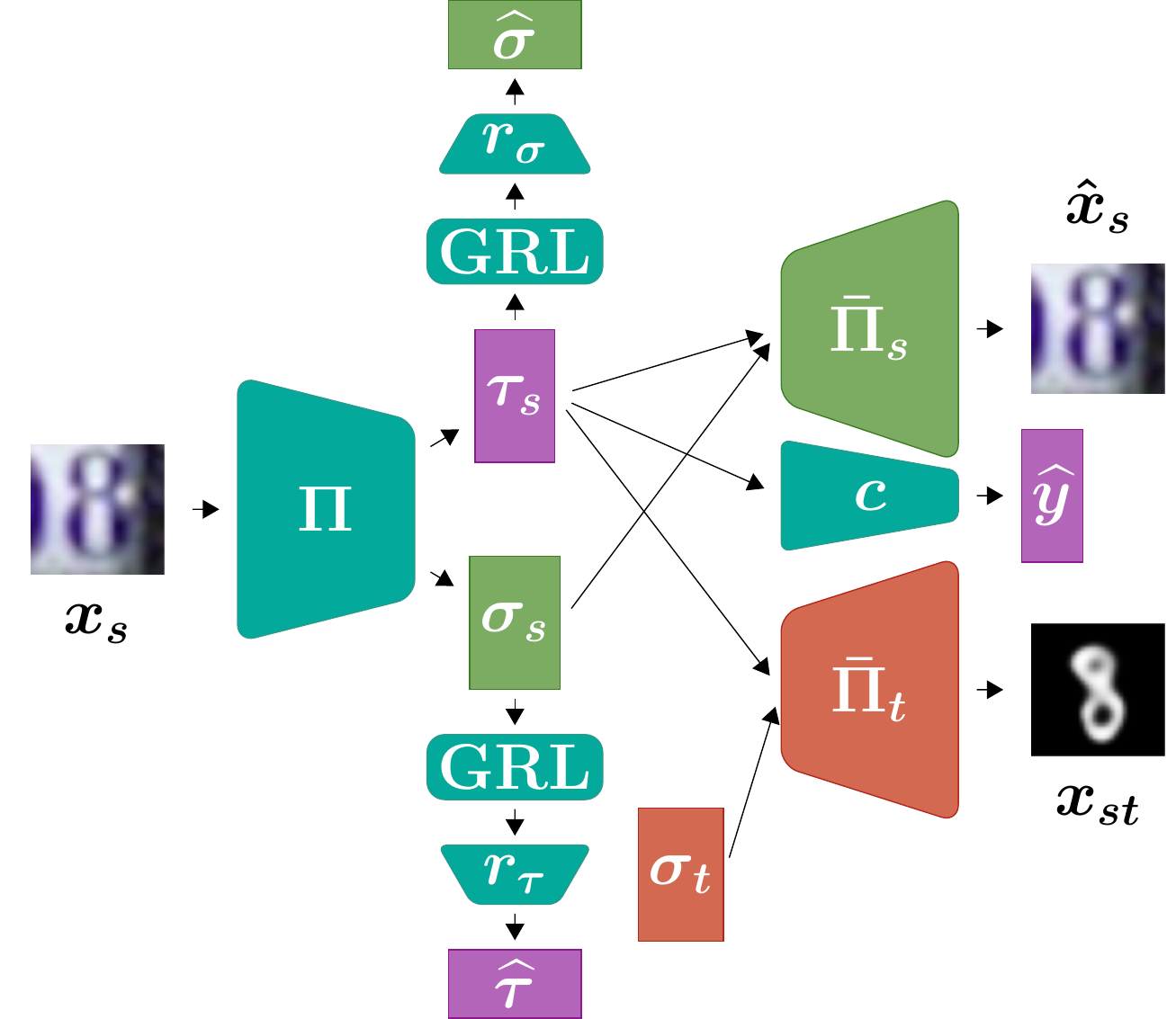}
    \end{center}
    \caption{Unsupervised domain adaptation}
    \label{fig:archi_adapt}
  \end{subfigure}
  \caption{Network architectures}
\label{fig:supervised_network}
\end{figure}

Our approach consists in estimating jointly $\Pi$, $\bar{\Pi}$ and $c$ as a deep feed-forward neural network. 
We shall note $\theta_\Pi$, $\theta_{\bar{\Pi}}$, and $\theta_c$ the parameters of the respective sub-parts of the network. 
$\bar{\Pi}\circ\Pi$ takes the form of an auto-encoder, while $c \circ \Pi_\tau$ is a task-related (classification or regression) network. 
Figure~\ref{fig:archi_supervised} summarizes the global architecture which we detail in the following paragraphs.


Conditions C1 and C2 are expressed through the definition of a task-specific loss $\mathcal{L}_{C1}$ (\emph{e.g.} cross-entropy for classification, L2 loss for regression) and a reconstruction loss $\mathcal{L}_{C2}$. 
Thus, the update of $\theta_\Pi$ should follow $-\nabla_{\theta_\Pi}\left(\mathcal{L}_{C1} + \mathcal{L}_{C2} \right)$, the update of $\theta_{\bar{\Pi}}$ relies on $-\nabla_{\theta_{\bar{\Pi}}}\mathcal{L}_{C2}$, and that of $\theta_c$ uses $-\nabla_{\theta_c} \mathcal{L}_{C1}$.

In order to achieve condition C3, we exploit Gradient Reversal Layers \cite[GRL]{ganin2016domain}.
Note that this choice is arbitrary and alternatives exist (e.g. \cite{sanchez2019learning}) to enforce C3.
We train two side networks $r_{\tau}: \mathcal{S} \rightarrow \mathcal{T}$ and $r_{\sigma}: \mathcal{T} \rightarrow \mathcal{S}$ whose purpose is to attempt to predict $T$ given $S$, and $S$ given $T$ respectively. 
For a given $x$, let us write $(\tau,\sigma)=\Pi(x)$,  $\widehat{\tau} = r_{\tau}(\sigma)$, and $\widehat{\sigma} = r_{\sigma}(\tau)$. 
We train $r_{\tau}$ and $r_{\sigma}$ to minimize the losses $\mathcal{L}_{r_\tau} =  \left \| \tau -  \widehat{\tau}\right \|_2$ and $\mathcal{L}_{r_\sigma} = \left \| \sigma -  \widehat{\sigma}\right \|_2$. 
Let $\mathcal{L}_{C3} = \mathcal{L}_{r_\tau} + \mathcal{L}_{r_\sigma}$ denote the combination of these losses. 
We connect these two sub-networks to the whole architecture using GRLs.
GRLs behave as the identity function during the forward pass and invert the gradient sign during the backward pass, hence pushing the parameters to maximize the output loss.
During training, this architecture constrains $\Pi$ to produce features in $\mathcal{T}$ and $\mathcal{S}$ with the least information shared between them. 
Consequently, the update of $\theta_\Pi$ follows $+\nabla_{\theta_\Pi}\mathcal{L}_{C3}$.

This constraint efficiently avoids information redundancy between $\mathcal{T}$ and $\mathcal{S}$. 
However, it does not avoid all the information being pushed into $\mathcal{T}$.
Preventing this undesirable behavior is the purpose of condition C4.
To translate C4 into a practical optimization loss, we consider a cyclic reconstruction scheme.
Consider two elements $x$ and $x'$ from $\mathcal{X}$, and their associated $(\tau,\sigma)=\Pi(x)$ and $(\tau',\sigma')=\Pi(x')$.
Let $\tilde{x}=\bar{\Pi}(\tau,\sigma')$ be the reconstruction of $\tau$ that uses the style $\sigma'$ of $x'$.
A correct allotment of the information between $\mathcal{T}$ and $\mathcal{S}$ requires that the task and style information be preserved in $(\tilde{\tau},\tilde{\sigma})=\Pi(\tilde{x})$.
In particular, we wish to have $\tilde{\sigma}$ as close as possible to $\sigma'$, and sufficiently far from $\sigma$.
We achieve this with a triplet loss \cite{schroff2015facenet} using $\tilde{\sigma}$ as the anchor, $\sigma'$ and $\sigma$ as, respectively, the positive and negative inputs, and a margin $m$. 
Note that alternatives, like regularization versus random projections (as e.g. in \cite{grill2020bootstrap,chen2020exploring}), might achieve a similar result without such a triplet loss.
Although it is not necessary for C4, we encourage the alignment of task-related representations by encouraging $\tilde{\tau}$ to be as close as possible to $\tau$, or, alternatively, to have $c(\tilde{\tau})$ as close as possible to $c(\tau)$.
Thus C4 results in minimizing the cyclic reconstruction loss $\mathcal{L}_{C4} =
    \left \| \tilde{\tau} -  \tau \right \|_2
    + max\{\left \| \tilde{\sigma} -  \sigma' \right \|_2
    - \left \| \tilde{\sigma} -  \sigma \right \|_2 + m , 0\}.$

The global loss enforcing disentanglement is thus $\mathcal{L}_{C1} + \mathcal{L}_{C2} - \mathcal{L}_{C3} + \mathcal{L}_{C4}$.
Specifically, with learning rate $\alpha$, the gradient-based update of network parameters boils down to:%
\begin{align*}
    \theta_{\Pi} & \leftarrow \theta_{\Pi} - \alpha \nabla_{\theta_{\Pi}} \left(\mathcal{L}_{C1} + \mathcal{L}_{C2} - \mathcal{L}_{C3} + \mathcal{L}_{C4} \right), \\
    \theta_{\bar{\Pi}} & \leftarrow \theta_{\bar{\Pi}} - \alpha \nabla_{\theta_{\bar{\Pi}}}  (\mathcal{L}_{C2} + \mathcal{L}_{C4}),\\
    \theta_{c} & \leftarrow \theta_{c} - \alpha \nabla_{\theta_c} \mathcal{L}_{C1},\\
    \theta_{r_{\tau}} & \leftarrow \theta_{r_{\tau}} - \alpha \nabla_{\theta_{r_{\tau}}} \mathcal{L}_{r_\tau}, \quad
    \theta_{r_{\sigma}} \leftarrow \theta_{r_{\sigma}} - \alpha \nabla_{\theta_{r_{\sigma}}}  \mathcal{L}_{r_\sigma}.
\end{align*}%
Computing these losses implies a forward pass for each sample in a minibatch, and a second forward pass to compute $\mathcal{L}_{C4}$. 
The backward propagation of gradients is unaffected. 
Thus, the computational complexity remains in the same class as vanilla empirical risk minimization.
We call this method DiCyR for Disentanglement by Cyclic Reconstruction.

\subsection{Task-style disentanglement in the unsupervised domain adaptation case}

We propose a variation of DiCyR for UDA, where we replace the decoder $\bar{\Pi}$ by two domain-specific decoders, $\bar{\Pi}_s$ and $\bar{\Pi}_t$. 
We shall compensate for the lack of labeled data in the target domain by computing cross-domain cyclic reconstructions. 

Let $(x_s,y_s)$ be a sample from the source domain and $x_t$ be a sample from the target domain. 
Let us denote $(\tau_s,\sigma_s)=\Pi(x_s)$ and $(\tau_t,\sigma_t)=\Pi(x_t)$, the corresponding projections in the latent task and style-related information spaces.
Then one can define, as in the previous section, $\mathcal{L}_{C1_s}$ as the task-specific loss on the source domain, and $\mathcal{L}_{C2_s}$ and $\mathcal{L}_{C2_t}$ as the reconstruction losses in the source and target domains respectively. 
As previously, we constrain the task-related representation and the style representation not to share information using two networks $r_{\tau}$ and $r_{\sigma}$, connected to the main architecture by GRL layers (Figure \ref{fig:archi_adapt}), allowing the definition of the $\mathcal{L}_{r_\tau}$, $\mathcal{L}_{r_\sigma}$ and $\mathcal{L}_{C3}$ losses. 
Lastly, we exploit cyclic reconstructions in both domains to correctly disentangle the information and hence define the same $\mathcal{L}_{C4}$ loss as above.
  
This disentanglement in the target domain separates the global information in two but does not guarantee that what is being pushed into $\tau$ is really the task-related information. 
The projection $\Pi_\tau$ could, for instance, retain confounding factors that are sufficient to classify inputs from the source domain, but not from the target domain, since $\mathcal{L}_{C1}$ is only defined for elements from the source domain.
This can only be enforced by cross-domain knowledge (since no correct labels are available in the target domain).
Thus, finally, we would like to allow projections from one domain into the other while retaining the task-related information, hence allowing domain adaption. 
Using the notations above, we construct $x_{st} = \bar{\Pi}_t(\tau_s,\sigma_t)$, the reconstruction of $x_s$'s task-related information, in the style of $x_t$. 
This creates an artificial sample in the target domain, whose label is $y_s$. 
Then, with $\left( \tau_{st},\sigma_{st} \right) = \Pi(x_{st})$, one wishes to have $\tau_{st}$ match closely $\tau_s$ (or, alternatively, $c(\tau_{st})$ match closely $y_s$) in order to prevent the loss of task information during the cross-domain projection
and thus to constrain the task representations to be domain-invariant.
Symmetrically, one can construct the artificial sample $x_{ts} = \bar{\Pi}_s(\tau_t,\sigma_s)$ and enforce that $\tau_{ts}$ closely matches $\tau_t$. 
Note that the label of $x_{ts}$ is unknown and yet it is still possible to enforce the disentanglement by cyclic reconstruction. 
Overall, these terms boil down to a cross-domain cyclic reconstruction loss for UDA $\mathcal{L}_{C1_t} = \left \| \tau_s - \tau_{ts}\right \|_2 + \left \| \tau_t - \tau_{st}\right \|_2$.

As previously, the global loss is the aggregate of all optimization criteria $\mathcal{L}_{C1_s}+\mathcal{L}_{C2_s}+\mathcal{L}_{C2_t}-\mathcal{L}_{C3}+\mathcal{L}_{C4}+\mathcal{L}_{C1_t}$.
We note $\mathcal{L}_{C1}=\mathcal{L}_{C1_s} + \mathcal{L}_{C1_t}$ and $\mathcal{L}_{C2}=\mathcal{L}_{C2_s} + \mathcal{L}_{C2_t}$ for brevity.
The network parameters are updated according to:%
\begin{align*}
    \theta_{\Pi} & \leftarrow \theta_{\Pi} - \alpha \nabla_{\theta_{\Pi}} (\mathcal{L}_{C1} + \mathcal{L}_{C2} - \mathcal{L}_{C3} + \mathcal{L}_{C4})\\
    \theta_{\bar{\Pi}_s} & \leftarrow \theta_{\bar{\Pi}_s} - \alpha \nabla_{\theta_{\bar{\Pi}_s}}  (\mathcal{L}_{C2_s} + \mathcal{L}_{C4})\\
    \theta_{\bar{\Pi}_t} & \leftarrow \theta_{\bar{\Pi}_t} - \alpha \nabla_{\theta_{\bar{\Pi}_t}}  (\mathcal{L}_{C2_t} + \mathcal{L}_{C4}),\\
    \theta_{c} & \leftarrow \theta_{c} - \alpha \nabla_{\theta_{c}} \mathcal{L}_{C1},\\
    \theta_{r_{\tau}} & \leftarrow \theta_{r_{\tau}} - \alpha \nabla_{\theta_{r_{\tau}}} \mathcal{L}_{r_\tau}\\
    \theta_{r_{\sigma}} & \leftarrow \theta_{r_{\sigma}} - \alpha \nabla_{\theta_{r_{\sigma}}}  \mathcal{L}_{r_\sigma}.
\end{align*}

\section{Experimental results and discussion}
\label{sec:xp}

We first evaluate DiCyR's ability to disentangle the task-related information from the style information in the supervised context. 
Then we demonstrate DiCyR's efficiency on UDA. 
Hyperparameters, network architectures and implementation choices are summarized in Appendix \ref{app:hyperparams}. 
We emphasize that no extensive hyperparameter tuning has been performed.

\subsection{Single-domain disentanglement}
\label{sec:xp-sup}

We evaluate the disentanglement performance of DiCyR by following the protocol introduced by \cite{mathieu2016disentangling}. 
Since we do not use generative models, we only focus on their two first items: \emph{swapping} and \emph{retrieval}. 
We evaluate DiCyR on the SVHN \cite{37648}, and 3D Shapes \cite{3dshapes18} disentanglement benchmarks. 
The task is predicting the central digit in the image for the SVHN dataset, and the shape of the central object in the scene for the 3D Shapes dataset.

\begin{figure}[h]
\centering
    \includegraphics[width=.4\linewidth]{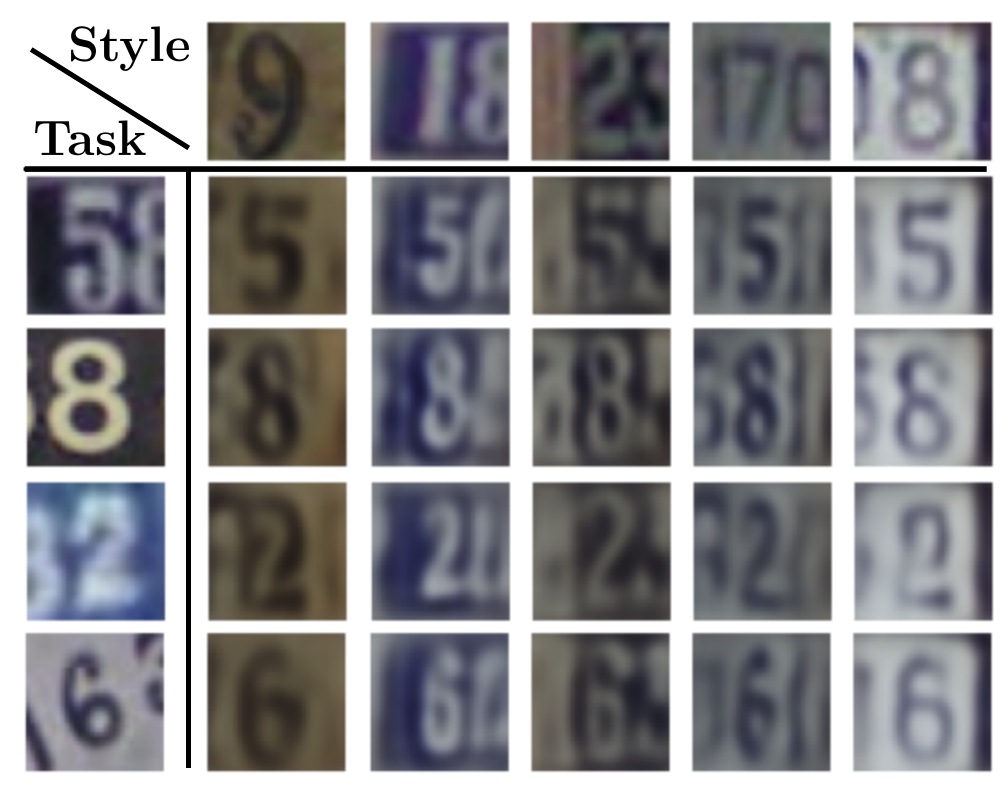}
    \label{fig:svhn}
    \includegraphics[width=.4\linewidth]{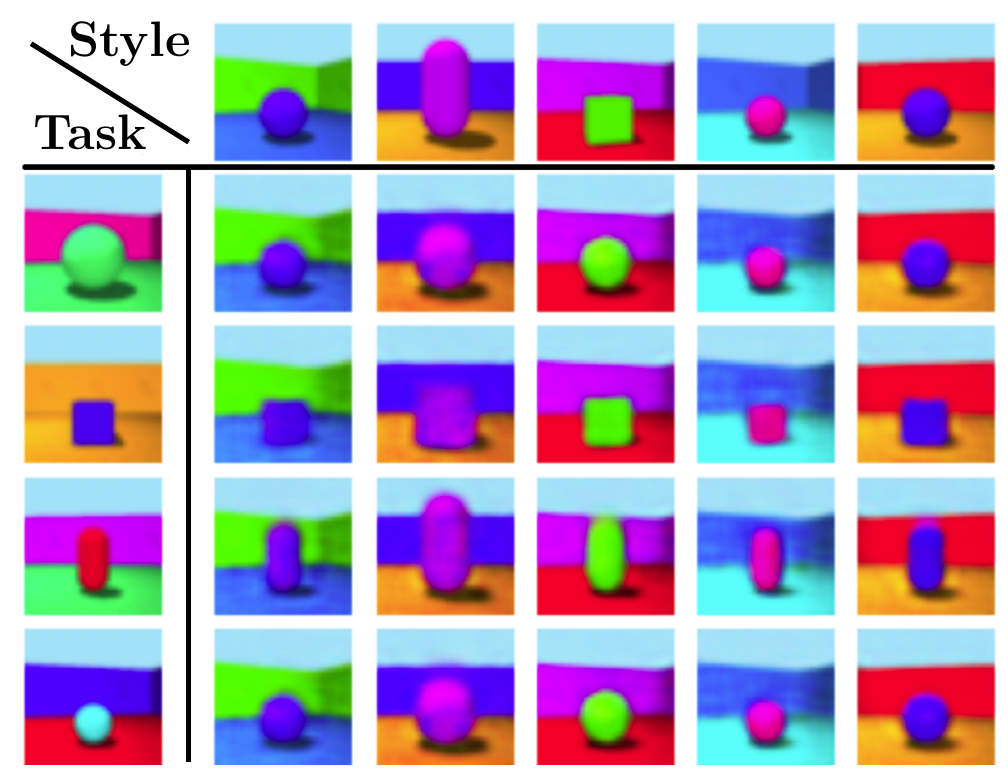}
    \label{fig:shapes}
\caption{Swapping styles on SVHN and 3D Shapes}
\label{fig:swapping}
\end{figure}

\emph{Swapping} involves swapping styles between samples and visually assessing the realism of the generated image. It combines the task-related information $\tau_i$ of a sample $x_i$ with the style $\sigma_j$ of another sample $x_j$. We use the decoder to produce an output $\tilde{x}_{ij}$. Figure \ref{fig:swapping} shows randomly generated outputs on the two datasets. DiCyR produces visually realistic artificial images with the desired styles.

\begin{figure}
\begin{subfigure}{\textwidth}
\centering
\includegraphics[width=.4\linewidth]{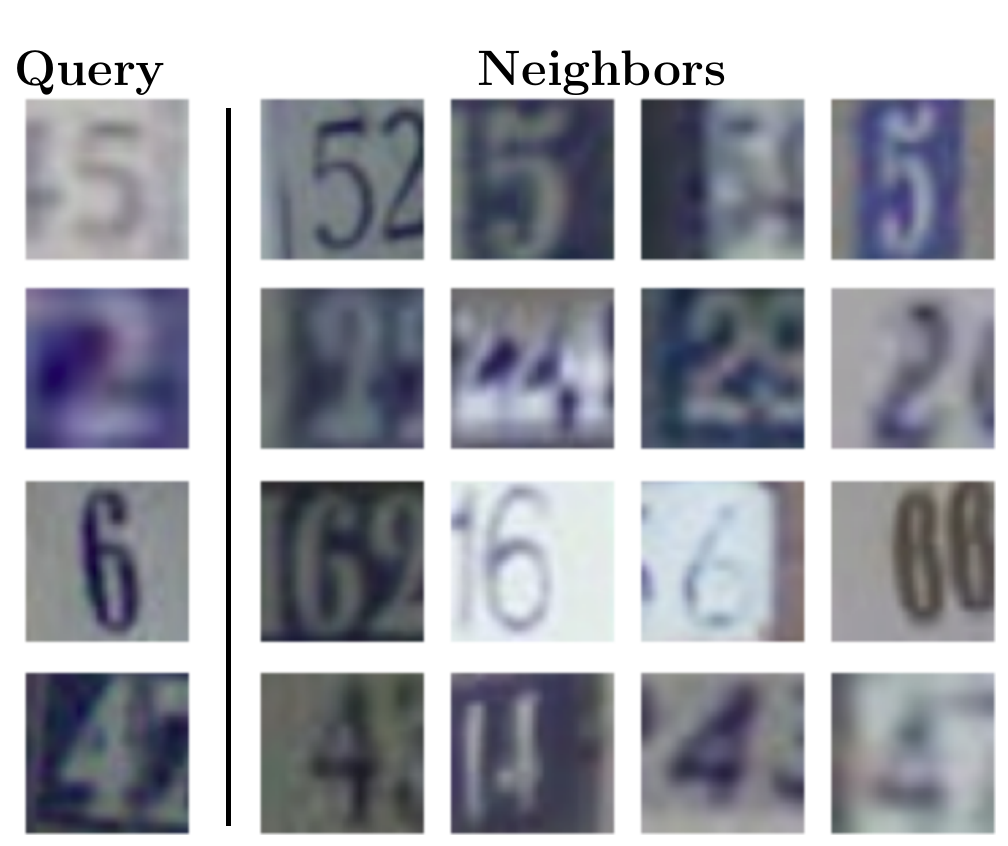} 
\includegraphics[width=.4\linewidth]{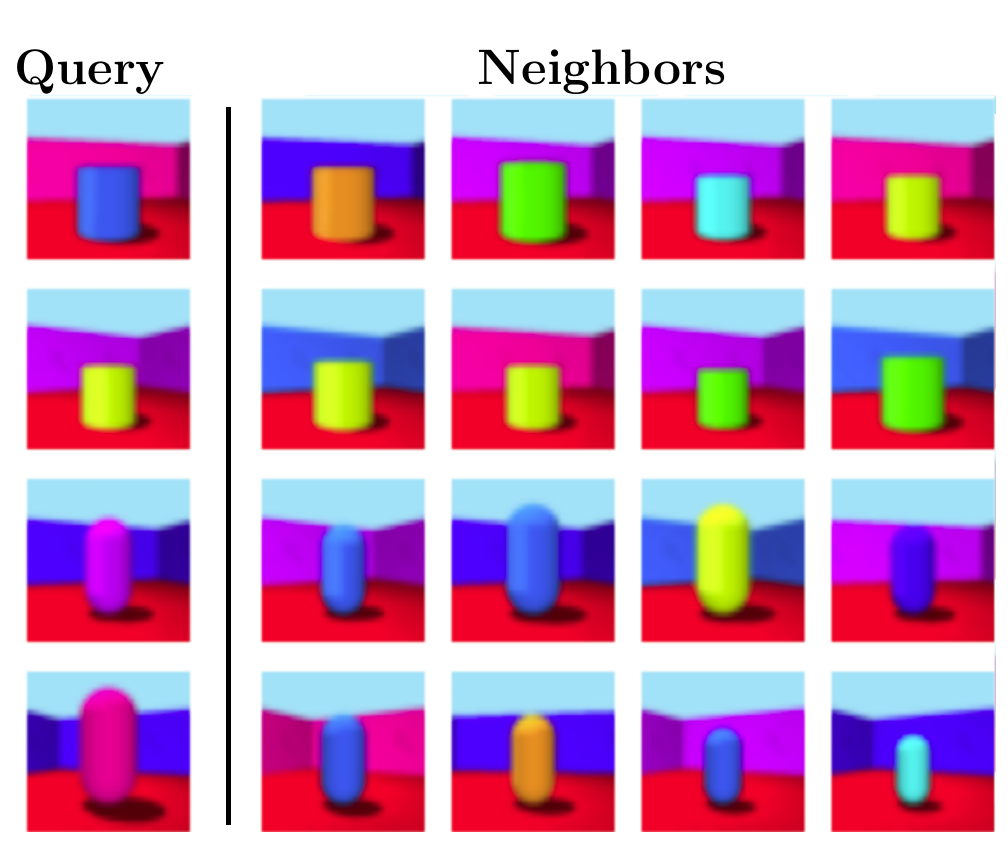}
\caption{Query on the task-related representation}
\end{subfigure}

\begin{subfigure}{\textwidth}
\centering
\includegraphics[width=.4\linewidth]{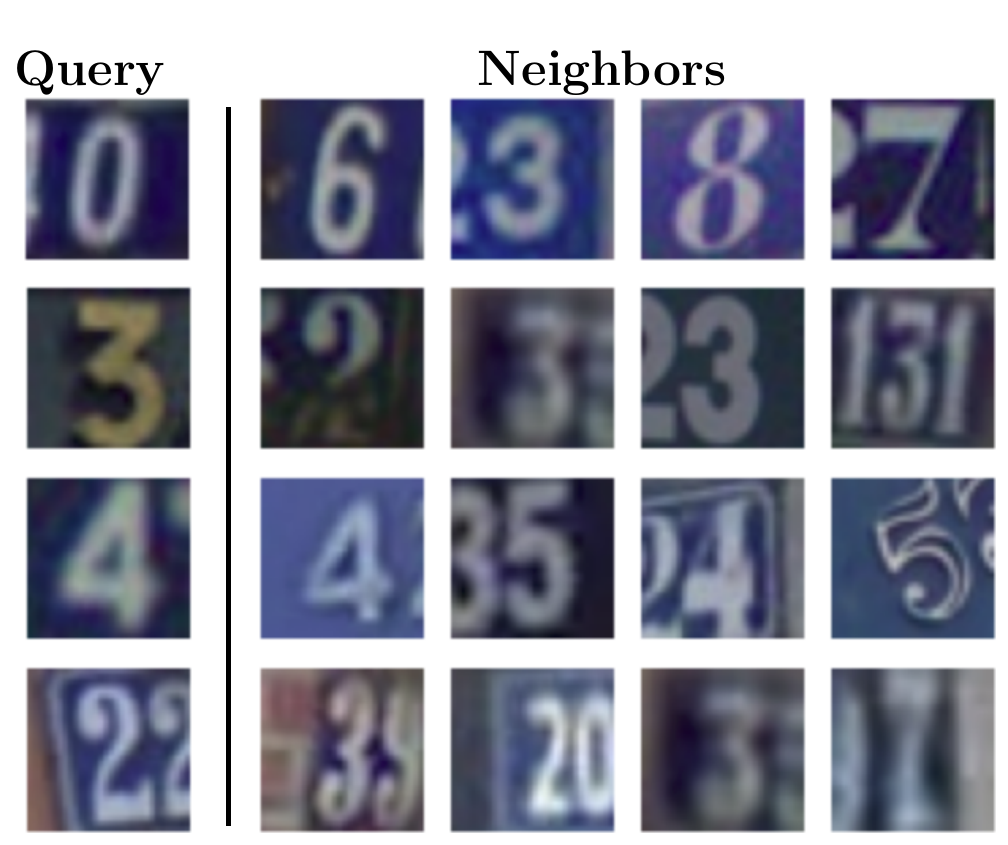} 
\includegraphics[width=.4\linewidth]{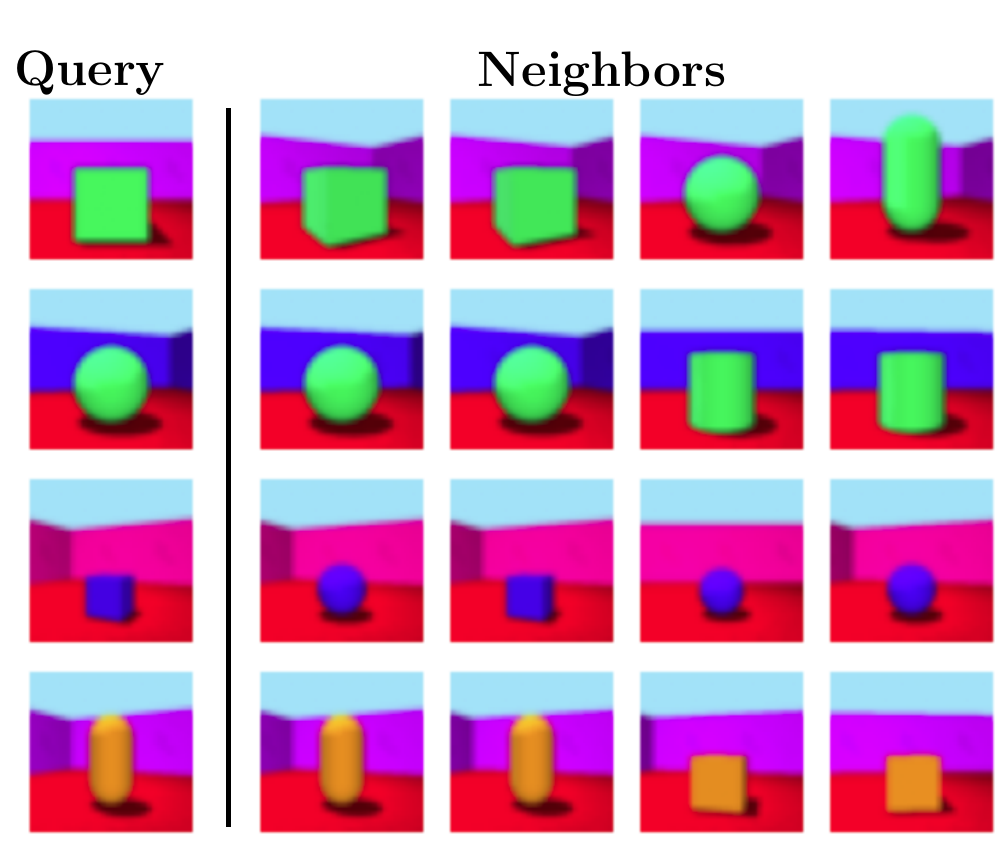}
\caption{Query on the style representation}
\end{subfigure}
\caption{Nearest neighbors according to each representation}
\label{fig:retrieval_task}
\end{figure}

\begin{table*}
\begin{center}
 \begin{tabular}{||c || c c || c c c c c||} 
 \hline
 Method & SVHN & 3D Shape & \makecell{floor\\hue} &  \makecell{wall\\hue} &  \makecell{object\\hue} & scale & orientation\\ 
 \hline\hline
 Full features & 0.98 & 1 & 0.94 & 0.94 & 0.89 & 0.6 & 0.5\\  
 \hline
 Task-related features& 0.98 & 1 & 0.11 & 0.12 & 0.13 & 0.15 & 0.10\\ 
 \hline
 Style features only & 0.17 & 0.26 & 0.89 & 0.95 & 0.88 & 0.59 & 0.42\\
 \hline
 Random guess & 0.10 & 0.25 & 0.10& 0.10 & 0.10 & 0.125 & 0.067\\
 \hline
\end{tabular}
\end{center}
 \caption{Classification accuracy using task, style or full information, on SVHN and 3D shapes.}
 \label{table:table1}
\end{table*}

\emph{Retrieval} concerns finding, in the dataset, the nearest neighbors in the embedding space for an image query. 
We carry out this search for nearest neighbors using the Euclidean distance on both the task-related and the style representations. 
A good indicator of the effectiveness of the information disentanglement would be to observe neighbors with the same labels as the query when computing distances on the task-related information space, and neighbors with similar style when using the style information. 
Figure \ref{fig:retrieval_task} demonstrate that the neighbors found when using the task-related information are samples with the same label as the query's label and that the neighbors found using the style representation share many characteristics with the query but not necessarily the same labels.

We ran a quantitative evaluation of disentanglement by training a neural network classifier with a single hidden layer of 32 units to predict labels, using either the task-related information alone, or the style information alone. 
If the information is correctly separated, we expect the classifier trained with task-related information only to get similar performance to a classifier trained with full information. 
Conversely, the classifier trained with the style information only should reach similar performance to a random guess (10\% accuracy on SVHN, 25\% on 3D Shapes). 
Table \ref{table:table1} (first two columns) reports the obtained testing accuracies.

It appears that the task-related representation contains enough information to correctly predict labels.
We also observe that full disentanglement is closely but not perfectly achieved, as the classifier trained only with style information behaves slightly better than random choice. 
To quantify how much style information is being unduly encoded in the task-related representation, we ran a similar experiment to predict the five other style variation factors in 3D Shapes (floor hue, wall hue, object hue, scale and orientation). The trained classifier reaches accuracies (Table \ref{table:table1} rightmost columns) that are very close to a random guess, thus validating the disentanglement quality.

\begin{figure}[h]
\centering
    \includegraphics[width=.4\linewidth]{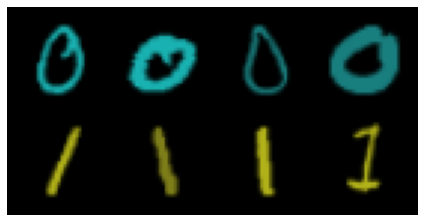}
    \label{fig:train}
    \includegraphics[width=.4\linewidth]{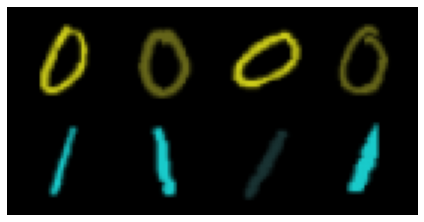}
    \label{fig:test}
\caption{Train and test data}
\label{fig:Bias}
\end{figure}

Without proper disentanglement, the features extracted by neural networks may contain context information that is specific to the training data distribution and unrelated to the task at hand.
This context information may comprehend confounding factors, introduced during the collection of the data for example, which can strongly affect generalization.
To illustrate this phenomenon, we train a network to classify zeros and ones extracted from the MNIST dataset. 
During training, we modify the dataset so that all ones have a yellow color and all zeros have a blue color while varying the intensity of the color. 
We then measure the accuracy of this network on a test dataset in which the colors of the ones and zeros are inverted (Figure \ref{fig:Bias}). The accuracies, reported in Table \ref{table:bias_results}, show that the network is incapable of generalizing on the test set. 
Conversely, by training DiCyR on the same training set, we observe that its accuracies remain similar despite the change of color. 
While the first network's predictions exploit the bias present in the training set's context information, DiCyR is insensitive to it.
Its ability to disentangle the task information from the context information allows DiCyR to retain its performance on data sampled from a different input distribution. 
Hence DiCyR helps generalize more robustly on unseen test distributions.

\begin{table}[h]
\begin{center}
 \begin{tabular}{||c || c | c ||} 
 \hline
 Method & Train accuracy & Test accuracy\\ 
 \hline\hline
 Supervised learning & 1.0 & 0.067\\  
 \hline
 DiCyR & 0.984 & 0.957\\
 \hline
\end{tabular}
\end{center}
 \caption{Vanilla classifier vs. DiCyR on a biased dataset}
 \label{table:bias_results}
\end{table}

\subsection{Unsupervised domain adaptation problem}
\label{sec:xp-usup}
We evaluate DiCyR by performing domain adaptation between the MNIST \cite{lecun1998gradient}, SVHN, and USPS \cite{hull1994database} datasets, and between the Syn-Signs \cite{ganin2015unsupervised} and the GTSRB \cite{stallkamp2011german} datasets.
Following common practice in the literature, we trained our network on four different settings: MNIST$\rightarrow$USPS, USPS$\rightarrow$MNIST, SVHN$\rightarrow$MNIST, and Syn-Signs$\rightarrow$GTSRB.
We measure the classification performance in the target domain and compare it with state-of-the-art methods (Table \ref{table:table2}).
We also compare with a baseline classifier that is only trained on the source domain data.
Values reported in Table \ref{table:table2} are quoted from their original papers.%
\footnote{
Comparisons might be inexact due to reproducibility concerns \cite{pineau2020improving} and these figures mostly indicate which are the top competing methods.}

\begin{table}[h]
\begin{center}
\begin{threeparttable}[t]

 \begin{tabular}{||l || l l l l||} 
 \hline
 \begin{tabular}{p{10mm} r}
                      & Source \\
                 \end{tabular}   & MNIST & USPS & SVHN & Syn-Signs\\ [0.5ex] 
 \begin{tabular}{p{10mm} r}
                   Method  & \hfill Target \\
                 \end{tabular}  & USPS & MNIST & MNIST & GTSRB\\ [0.5ex] 
 \hline\hline
 Baseline & $78.1$ & $58.0$ & $60.2$ & $79.0$\\
 \hline
 DSN \cite{bousmalis2016domain} & $91.3$ & - & $82.7$  & $93.1$\\
 \hline
 DiDA \cite{cao2018dida} & $92.5$ & - & $83.6$  & -\\
 \hline
 SBADA-GAN \cite{russo2018source} & $97.6$ & $95.0$ & $76.1$ & $96.7$\\
 \hline
 CyCADA \cite{hoffman2018cycada} & $95.6$ & $96.5$ & $90.4$ & -  \\
 \hline
 ACAL \cite{hosseini-asl2018augmented} & $98.3$ & $97.2$ & $96.5$ & -\\
 \hline
 DiCyR (ours) & \textbf{98.7} & \textbf{98.3} &  \textbf{97.7} & \textbf{97.4}\\
 \hline\hline
 DANN \cite{ganin2016domain} &  $85.1$ & $73.0$ & $73.9$ & $88.6$\\ 
 \hline
 ADDA \cite{tzeng2017adversarial} & $89.4$ & $90.1$ & $76.0$  & -\\
 \hline
 DRCN \cite{ghifary2016deep} & $91.8$ & $73.7$ & $82.0$ & -\\
 \hline
 DWT \cite{roy2019unsupervised} & $99.1$ & $98.8$ & $97.7$ & - \\
 \hline
 SEDA \cite{french2018selfensembling} & $98.2$ & $99.5$ & $99.3$ & 99.3\\
 \hline
 
 SHOT \cite{liang2020we} & $98.4$ & $98.0$ & $98.9$ & -  \\
 \hline
\end{tabular}
 \footnotesize
 \centering

 \end{threeparttable}
  \end{center}
 \caption{Target domain accuracy, reported as percentages}
 \label{table:table2}
\end{table}

Our method, without extensive hyperparameter tuning, appears to be on par with the best state-of-the-art methods. 
We separate the methods that perform domain adaptation between those that aim at disentanglement or image-to-image translation (DSN, DiDA, SBADA-GAN, CyCADA, ACAL), and those that rely on other principles such as features or domain statistics alignment (DANN, ADDA, DRCN, DWT, SEDA, SHOT).
DiCyR fundamentally belongs to the first group and outperforms all methods therein, on all benchmarks.
We underline that this is probably the key result of this contribution since DiCyR is primarily a disentanglement method (which we apply to domain adaptation, among other tasks).

DiCyR is only slightly outmatched by DWT and SEDA on the MNIST$\leftrightarrow$USPS and by SEDA and SHOT in the SVHN$\rightarrow$MNIST benchmarks.
The variation on batch normalization introduced by DWT and the mean teacher semi-supervised learning model \cite{tarvainen2017mean} used by SEDA are orthogonal to our contribution and could be combined to DiCyR in order to improve its performance.

DiCyR uses GRLs to ensure that no information is shared between $\mathcal{T}$ and $\mathcal{S}$. 
One might object that condition C3 was expressed in terms of mutual information.
Thus, DiCyR only indirectly implements this condition using GRLs.
An alternative could be to use an estimator of the mutual information, such as proposed by \cite{belghazi2018mutual}, to directly minimize it (and thus avoid the adversarial setting altogether).
Such an approach was explored in the work of \cite{sanchez2019learning} to disentangle representations between pairs of images, and constitutes a promising perspective of research.

\begin{figure}[H]
\centering
\begin{subfigure}{\textwidth}
\includegraphics[width=.49\linewidth]{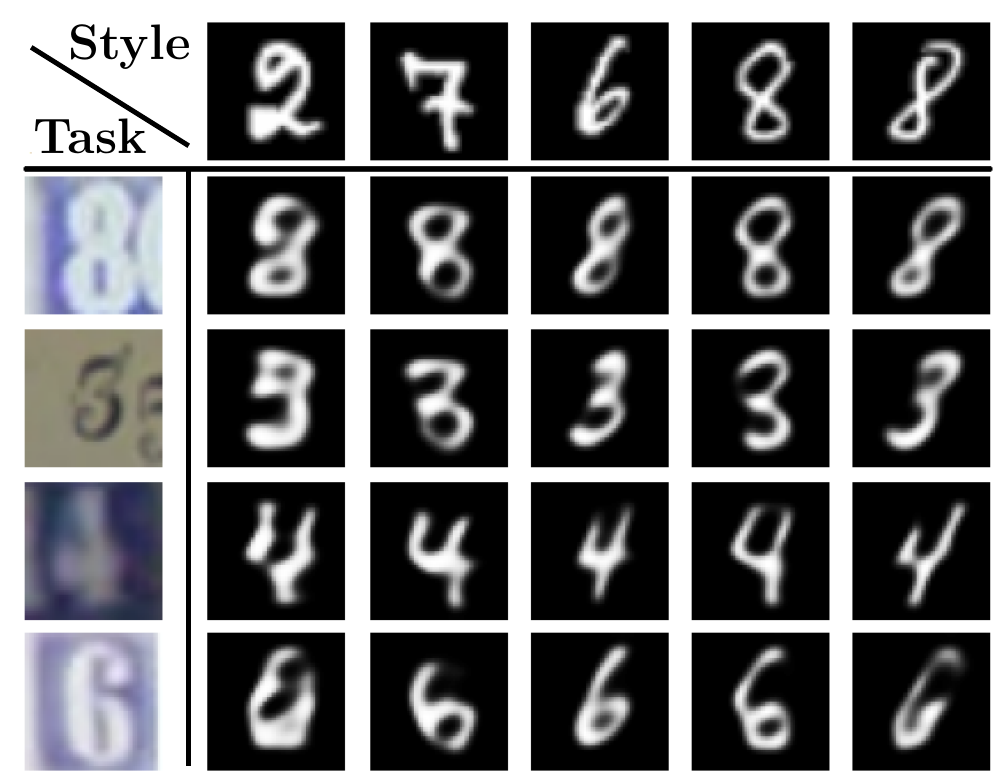} 
\includegraphics[width=.49\linewidth]{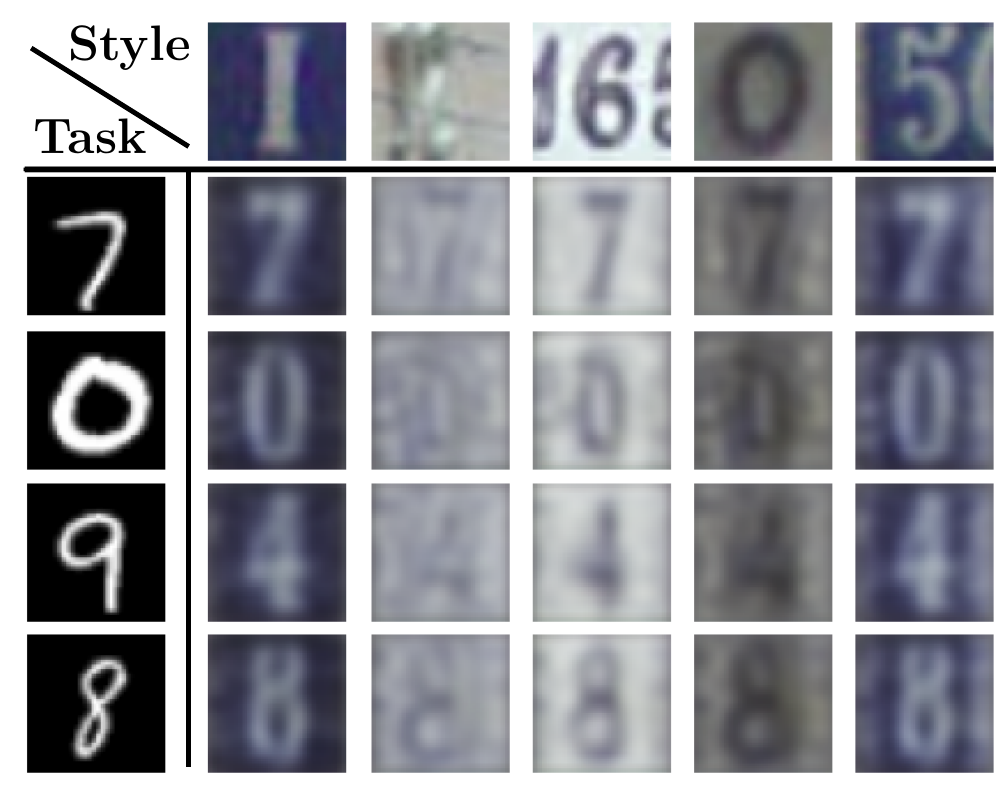}
\caption{Swapping between SVHN (source) and MNIST (target)}
\end{subfigure}

\begin{subfigure}{\textwidth}
\includegraphics[width=.49\linewidth]{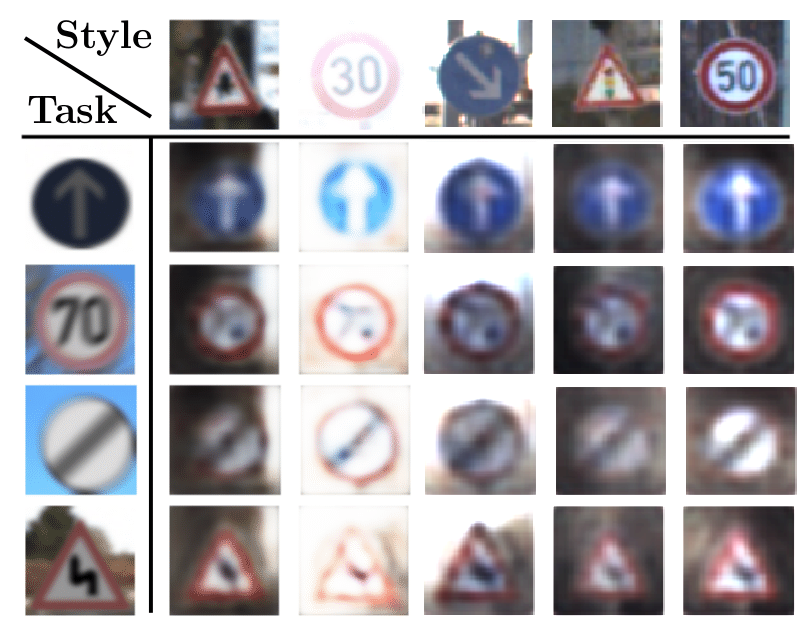} 
\includegraphics[width=.49\linewidth]{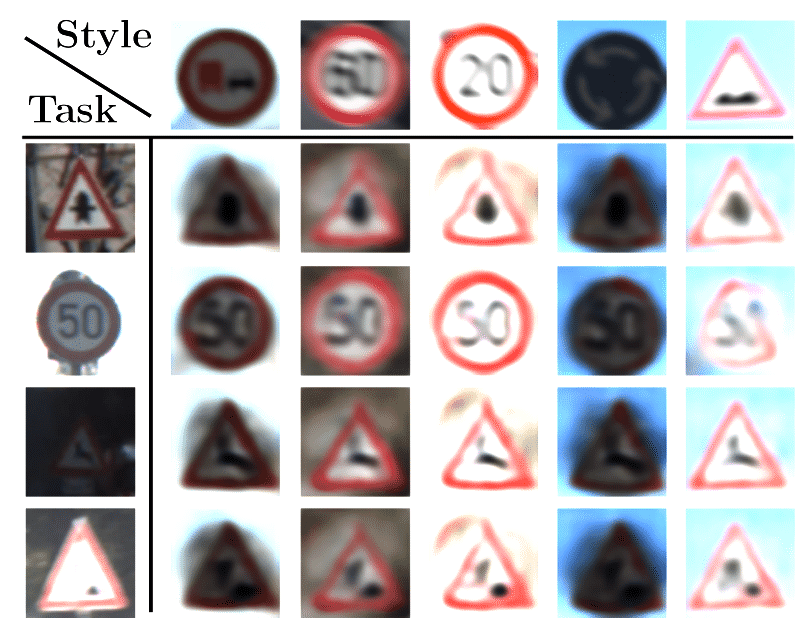}
\caption{Swapping between Syn-Signs (source) and GTSRB  (target)}
\end{subfigure}
\caption{Cross-domain swapping}
\label{fig:cross_domain}
\end{figure}

\begin{figure}[H]
\centering
\begin{subfigure}{\textwidth}
\includegraphics[width=.49\linewidth]{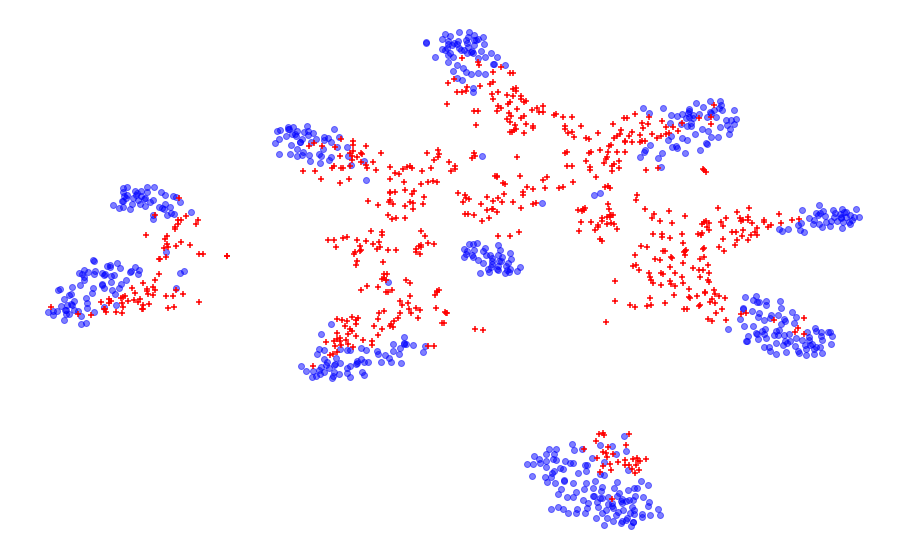} 
\includegraphics[width=.49\linewidth]{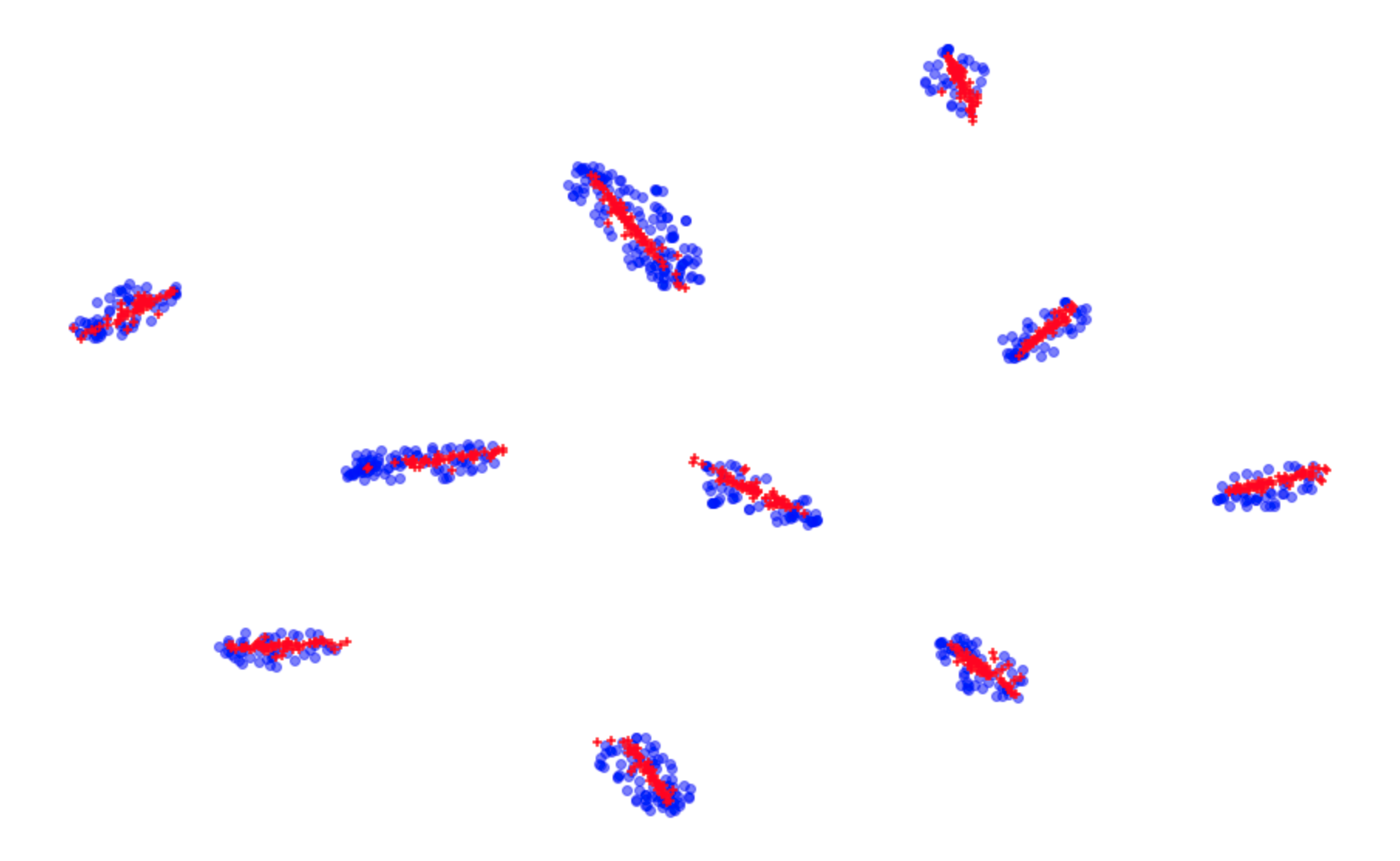}
\end{subfigure}
\caption{t-SNE on task-related features. SVHN (blue) $\rightarrow$ MNIST (red). Left: source only. Right: DiCyR}
\label{fig:tsne}
\end{figure}

A desirable property of the task-related encoding is its domain invariance. To evaluate this aspect, we built a t-SNE representation \cite{hinton2003stochastic} of the task-related features, in order to verify their alignment between domains (Figure \ref{fig:tsne}).

As in Section \ref{sec:xp-sup}, we evaluate qualitatively the effectiveness of disentanglement, especially in the target domain, and produce visualizations of cross-domain style and task swapping. 
Here, we combine one domain's task information with the other domain's styles to reconstruct the images of Figure \ref{fig:cross_domain}.
The most important finding is that the style information was correctly disentangled from the task-related information in the target domain without the use of any label. 
Specifically, the rows in these figures show that the class information is preserved when a new style is applied, while the columns illustrate the efficient style transfer allowed by disentanglement. 

\looseness=-1
Finally, directly computing the distances on the task-related features in $\mathcal{L}_{C1_t}$ often leads to unstable results.
As hinted in Section \ref{sec:contrib}, using instead a task oriented loss $\mathcal{L}_{C1_t} = \left \| c(\tau_s) - y\right \|_2 + \left \| c(\tau_t) - c(\tau_{st}) \right \|_2$ stabilizes training and improves the target domain accuracy.
Training $c$ with cross-domain projections from the source domain and the corresponding labels improves its generalization to the target domain and forces the encoder to produce task-related features common to both domains.
To illustrate this property, consider the following example. 
In one domain, the digit ``7'' is written with a middle bar, while in the other it has none. 
This domain-specific middle bar feature should not be expressed in $\mathcal{T}$; it should be considered as a task-orthogonal style feature.
Thus using $c$'s predictions within the domain cyclic loss, instead of distances in $\mathcal{T}$, prevents the encoder from representing the domain-specific features in $\mathcal{T}$ and encourages their embedding in $\mathcal{S}$.
\section{Conclusion}
\label{sec:conclusion}

In this work, we introduced a new disentanglement method, called DiCyR, to separate task-related and task-orthogonal style information into different representations in the context of unsupervised domain adaptation. 
This method also provides a simple and efficient way to obtain disentangled representations for supervised learning problems. 
Its main features are its overall simplicity, the use of intra-domain and cross-domain cyclic reconstruction, and information separation through Gradient Reversal Layers. 
The design of this method stems from a formal definition of disentanglement for domain adaptation which, to the best of our knowledge, is new. 
Empirical evaluation shows that DiCyR allows for efficient disentanglement, as demonstrated on both information retrieval and domain adaptation tasks where it is competitive with state-of-the-art methods. Moreover, it is the only method that explicitly aims at disentanglement in the target domain, where no label information is available.
%
%
%

%
%
\section*{Acknowledgments}
The authors acknowledge the support of the DEEL project, the funding of the AI Interdisciplinary Institute ANITI
funding, through the French “Investing for the Future – PIA3” program under grant agreement
ANR-19-PI3A-0004.
They would also like to thank Dennis Wilson and Thomas Oberlin for fruitful discussions and comments on the paper.


\newpage
\appendix

\section{Cross-domain disentanglement visualizations}
\label{app:cross_dom_disentangle}

\begin{figure}[h]
\centering
    \includegraphics[width=.48\linewidth]{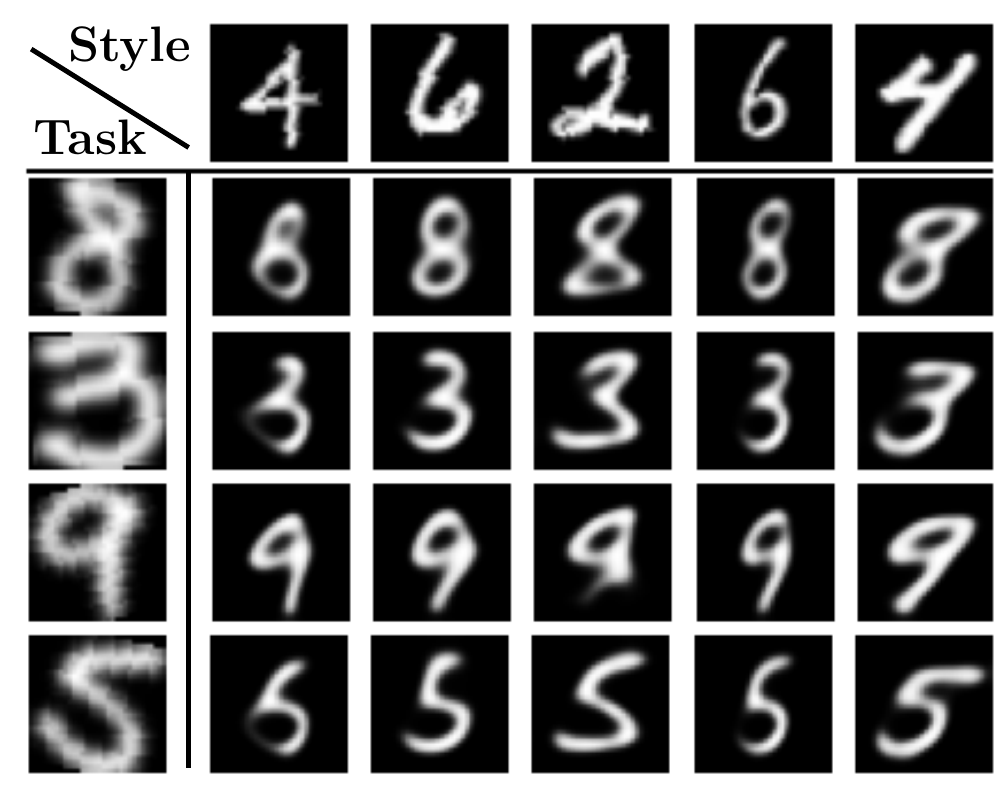}
    \label{fig:mnist_cross1}
    \includegraphics[width=.48\linewidth]{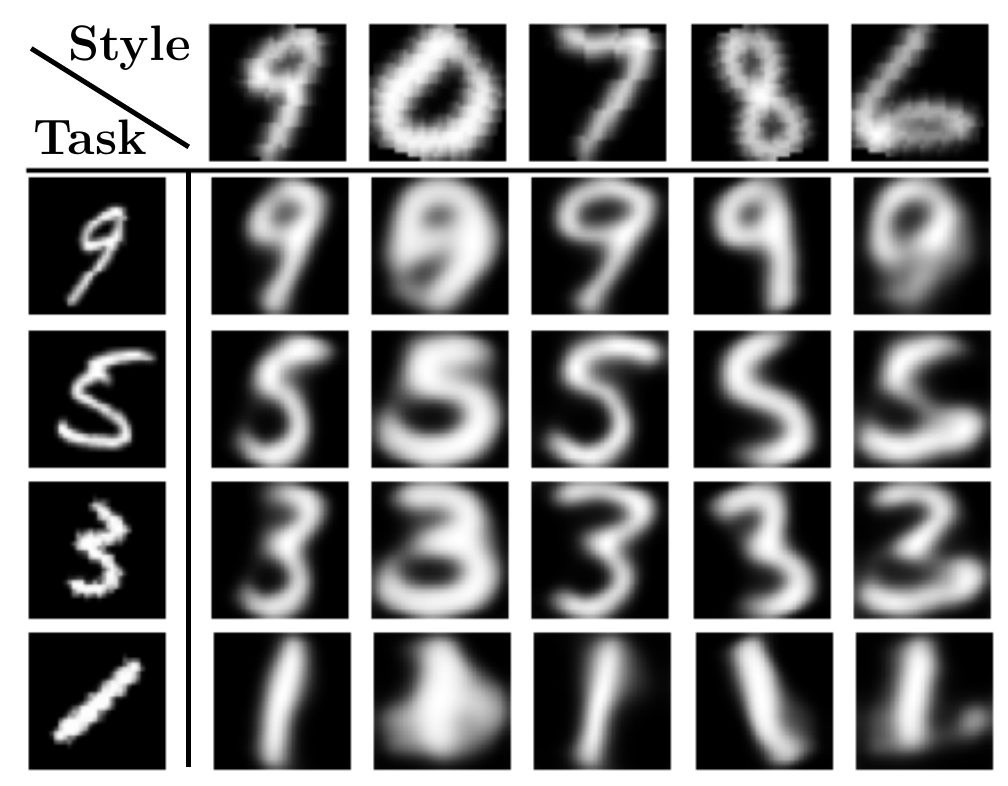}
    \label{fig:mnist_cross2}
\caption{Cross-domain swapping between USPS (source) and MNIST (target)}
\label{fig:swapping_mnist}
\end{figure}

\begin{figure}[h]
\centering
    \includegraphics[width=.48\linewidth]{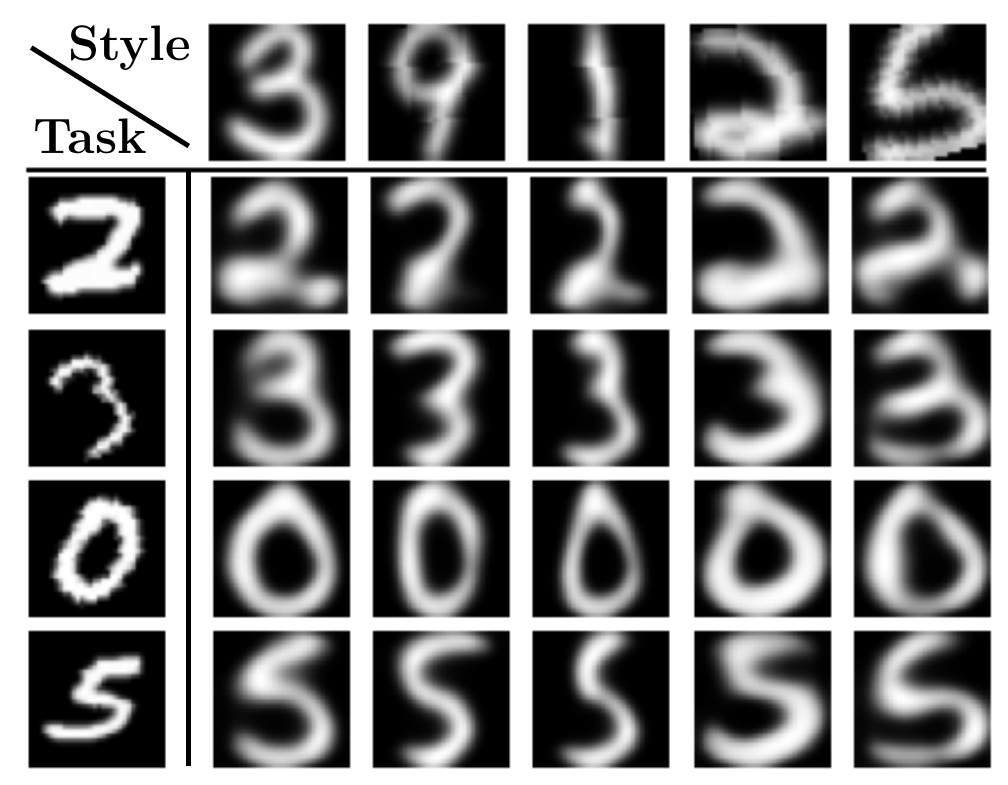}
    \label{fig:usps_cross1}
    \includegraphics[width=.48\linewidth]{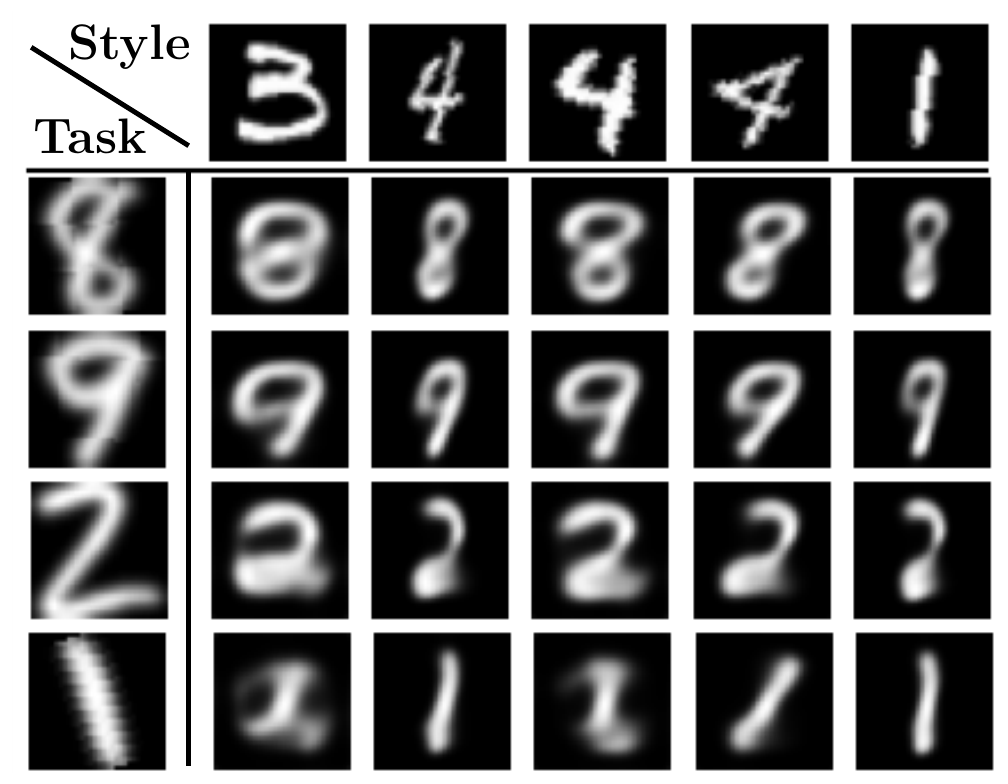}
\caption{Cross-domain swapping between MNIST (source) and USPS (target)}
\label{fig:swapping_usps}
\end{figure}

Figures \ref{fig:swapping_mnist} and \ref{fig:swapping_usps} report extra cross-domain visualizations similar to those in Figure \ref{fig:cross_domain}.

\section{Discussion on the $\beta_{C3}$ and $\beta_{C1_t}$ scheduling}
\label{app:schedule}

Although the schedule on $\beta_{C3}$ (single domain case) and $\beta_{C1_t}$ (domain adaptation case) is not absolutely necessary, we found out it helped the overall convergence. These coefficients gradually increase the weight of the information disentanglement objective and the cross-domain reconstruction objective. This assigns more importance to learning a good predictor $c \circ \Pi$ during early stages. From this perspective, gradually increasing $\beta_{C3}$ can be seen as gradually removing task-useless information from $\mathcal{T}$ and transferring it to $\mathcal{S}$. Similarly, increasing $\beta_{C1_t}$ corresponds to letting the network discover disentangled representations before aligning them across domains.

As previously mentioned, our goal in this study was to provide a robust disentanglement method that permits domain adaptation. Therefore, no complete hyper-parameter study and tuning was performed and these findings are thus reported as such and might be incomplete. Refining the understanding of the influence of the different $\beta$ coefficients is closer to the problem of meta-learning and beyond the scope of this paper.
\section{Influence of batch normalization and dropout}

Batch normalization \cite{ioffe2015batch} is an efficient way to reduce the discrepancy between the source and target distributions statistics.
We noticed that, for the specific SVHN $\rightarrow$ MNIST setting, using instance normalization \cite{ulyanov2016instance} slightly improves the target domain accuracy.
Normalizing across channels, the instance normalization layers helps the networks to be agnostic to the image contrast which is particularly strong in MNIST.
We also noticed that using a large dropout in the sub-network $c$, and small embedding dimensions for $\Pi$'s outputs improves both the disentanglement quality and the target domain accuracy.
We conjecture that the induced information bottleneck forces the task-related representation to be as concise as possible and thus encourages disentanglement.

\section{Network architecture and hyper-parameters}
\label{app:hyperparams}

In line with the Machine Learning Reproducibility Checklist \cite{pineau2020improving}, we provide all details about network architectures, training hyperparameters and computing resources to permit easy reproduction of our results. 
The next paragraphs detail the network architectures used in the experiments of Section \ref{sec:xp}. It should be noted that neither these architectures, nor the associated hyper-parameters have been extensively and finely tuned to their respective tasks, as the goal of this contribution was to provide a generic, robust method. Thus, it is likely that performance gains can still be obtained on this front.

\subsection{Single domain supervised disentanglement experiments}
\label{app:Supervisedhyperparams}
This section and Table \ref{tab:supervised} describe the network architecture and the hyper-parameters used in the experiments of Section \ref{sec:xp-sup}. 
The encoder $\Pi$ is composed of shared layers, followed by the specific single-layer task-related and style encodings $\Pi_\tau$ and $\Pi_\sigma$. 
We used the exact same network architectures for both the 3D shapes and SVHN datasets, the only difference being the dimension of the embeddings $\mathcal{T}$ and $\mathcal{S}$.

In all experiments, we found out empirically that convergence was improved by applying a coefficient $\beta_{C2}=5$ to $\mathcal{L}_{C2}$ and $\beta_{C4}=0.1$ to $\mathcal{L}_{C4}$ in the global loss. 
We also use a $\beta_{C3}$ on $\mathcal{L}_{C3}$; this coefficient increases linearly from $10^{-2}$ to $10$ over the first 10 epochs and remains at 10 afterwards (see Appendix \ref{app:schedule} for a discussion on this coefficient).
Convergence was reached within 50 epochs.
We used Adam \cite{kingma2015adam} as an optimizer with a learning rate $lr=5 \cdot 10^{-4}$ for the first 30 epochs and $lr=5 \cdot 10^{-5}$ for the last 20 epochs.

\begin{table}[h]
  \begin{center}
    \begin{threeparttable}[t]
      \begin{tabular}{||c | l||}
        \hline
        Network                                    & Architecture                                                                         \\
        \hline

        \multirow{6}{1em}{$\Pi$}                   & $\bullet $ Conv2D(filters=32, kernel=$5\times5$, stride=1, padding=2, ReLU)          \\
                                                   & $\bullet $ Max Pooling(filters=$2\times2$, stride=2)                                 \\
                                                   & $\bullet $ Conv2D(filters=32, kernel=$5\times5$, stride=1, padding=2, ReLU)          \\
                                                   & $\bullet $ Max Pooling(filters=$2\times2$, stride=2)                                 \\
                                                   & $\bullet $ Conv2D(filters=64 for SVHN, kernel=$3\times3$, stride=1, padding=1, ReLU) \\
                                                   & $\bullet $ Dense(nb\_neurons=1024, ReLU)                                             \\
        \hline
        \hline

        $\Pi_\tau$                                 & $\bullet $ Dense(nb\_neurons=150, ReLU)                                              \\
        \hline
        \hline

        $\Pi_\sigma$                               & $\bullet $ Dense(nb\_neurons=150, ReLU)                                              \\
        \hline
        \hline

        \multirow{7}{1em}{$\bar{\Pi}$}             & $\bullet $ Dense(nb\_neurons=1024, ReLU)                                             \\
                                                   & $\bullet $ Dense(nb\_neurons=8192, ReLU)                                             \\
                                                   & $\bullet $ Conv2D(filters=64, kernel=$3\times3$, stride=1, padding=1, ReLU)          \\
                                                   & $\bullet $ Upsample (scale\_factor=2)                                                \\
                                                   & $\bullet $ Conv2D(filters=32, kernel=$5\times5$, stride=1, padding=2, ReLU)          \\
                                                   & $\bullet $ Upsample(scale\_factor=2)                                                 \\
                                                   & $\bullet $ Conv2D (filters=3, kernel=$5\times5$, stride=1, padding=2, Sigmoid)       \\
        \hline
        \hline

        \multirow{2}{1em}{$c$}                     & $\bullet $ Dropout(p=0.2(3DShapes) or 0.55(SVHN))                                    \\
                                                   & $\bullet $ Dense(nb\_neurons=nb\_labels, Softmax)                                    \\
        \hline
        \hline

        \multirow{2}{5em}{$r_\tau$ and $r_\sigma$} & $\bullet $ Gradient Reversal Layer                                                   \\
                                                   & $\bullet $ Dense(nb\_neurons=100, ReLU)                                              \\
                                                   & $\bullet $ Dense(nb\_neurons=100, ReLU)                                              \\
                                                   & $\bullet $ Dense(nb\_neurons=20(Shapes) or 150(SVHN), Linear)                        \\
        \hline
      \end{tabular}
    \end{threeparttable}
  \end{center}
  \caption{Supervised experiments architectures}
  \label{tab:supervised}
\end{table}

\subsection{Unsupervised domain adaptation experiments}
\label{app:UDAhyperparams}

This section and Tables \ref{fig:archi_DA1},  \ref{fig:archi_DA2}, and \ref{fig:archi_DA3} describe the network architecture and the hyper-parameters used in the experiments of Section \ref{sec:xp-usup}. 
The encoder $\Pi$ is composed of shared layers, followed by the specific task-related and style encodings. 
Those final layers are denoted $\Pi_\tau$ and $\Pi_\sigma$ in the tables below.
For the sake of implementation simplicity, we chose to project samples from the source domain and samples from the target domain into two separate style embeddings (one for each domain). 
Thus $\Pi_\sigma$ is actually duplicated in two heads $\Pi_{\sigma, s}$ and $\Pi_{\sigma, t}$ with the same structure and output space.
In all experiments, 
we applied a coefficient $\beta_{C4}=0.1$ to $\mathcal{L}_{C4}$ and $\beta_{C1_t}$ to $\mathcal{L}_{C1_t}$, with $\beta_{C1_t}$ increasing linearly from 0 to 10 during the 10 first epochs and remaining at 10 afterwards (see Section \ref{app:schedule} for a discussion). 
Convergence was reached within 50 epochs (generally within 30 epochs). 
We used Adam \cite{kingma2015adam} as an optimizer with a learning rate $lr=5 \cdot 10^{-4}$ for the first 30 epochs and $lr=5 \cdot 10^{-5}$ for the last 20 epochs.

\begin{table}[h]
\small
  \begin{center}
      \begin{tabular}{||c | l||}
        \hline
        Network                                            & Architecture                                                                                            \\
        \hline

        \multirow{11}{1em}{$\Pi$}                          & $\bullet $ Conv2D(filters=32, kernel=5×5, stride=1, padding=2, Linear)                                  \\
                                                           & $\bullet $ Instance Normalization                                                                       \\
                                                           & $\bullet $ Max Pooling(filters=$2\times2$, stride=2)                                                    \\
                                                           & $\bullet $ Conv2D(filters=32(SVHN$\rightarrow$MNIST) or 64(MNIST$\rightarrow$SVHN), kernel=$5\times5$,  \\
                                                           & $\space$  stride=1, padding=2, Linear)                                                                  \\
                                                           & $\bullet $ Instance Normalization                                                                       \\
                                                           & $\bullet $ Max Pooling(filters=$2\times2$, stride=2)                                                    \\
                                                           & $\bullet $ Conv2D(filters=32(SVHN$\rightarrow$MNIST) or 128(MNIST$\rightarrow$SVHN), kernel=$3\times3$, \\
                                                           & $\space$  stride=1, padding=2, Linear)                                                                  \\
                                                           & $\bullet $ Instance Normalization                                                                       \\
                                                           & $\bullet $ Dense(nb\_neurons=1024, ReLU)                                                                \\
        \hline
        \hline

        $\Pi_\tau$                                         & $\bullet $ Dense(nb\_neurons=75(SVHN$\rightarrow$MNIST) or 200(MNIST$\rightarrow$SVHN), ReLU)           \\
        \hline
        \hline

        $\Pi_\sigma$                                       & $\bullet $ Dense(nb\_neurons=75(SVHN$\rightarrow$MNIST) or 200(MNIST$\rightarrow$SVHN), ReLU)           \\
        \hline
        \hline

        \multirow{7}{5em}{$\bar{\Pi}_s$ and $\bar{\Pi}_t$} & $\bullet $ Dense(nb\_neurons=1024, ReLU)                                                                \\
                                                           & $\bullet $ Dense(nb\_neurons=2048, ReLU)                                                                \\
                                                           & $\bullet $ Conv2D(filters=32, kernel=$3\times3$, stride=1, padding=1, ReLU)                             \\
                                                           & $\bullet $ Upsample (scale\_factor=2)                                                                   \\
                                                           & $\bullet $ Conv2D(filters=32, kernel=$5\times5$, stride=1, padding=2, ReLU)                             \\
                                                           & $\bullet $ Upsample(scale\_factor=2)                                                                    \\
                                                           & $\bullet $ Conv2D (filters=3, kernel=$5\times5$, stride=1, padding=2, Sigmoid)                          \\
        \hline
        \hline

        \multirow{2}{1em}{$c$}                             & $\bullet $ Dropout(p=0.55)                                                                              \\
                                                           & $\bullet $ Dense(nb\_neurons=10, Softmax)                                                               \\
        \hline
        \hline

        \multirow{3}{5em}{$r_\tau$ and $r_\sigma$}         & $\bullet $ Gradient Reversal Layer                                                                      \\
                                                           & $\bullet $ Dense(nb\_neurons=100, ReLU)                                                                 \\
                                                           & $\bullet $ Dense(nb\_neurons=75(SVHN$\rightarrow$MNIST) or 200(MNIST$\rightarrow$SVHN), Linear)         \\
        \hline
      \end{tabular}
  \end{center}
  \caption{SVHN$\leftrightarrow$MNIST networks architectures}
  \label{fig:archi_DA1}
\end{table}

\begin{table}[h]
\small
  \begin{center}
      \begin{tabular}{||c | l||}
        \hline
        Network                                            & Architecture                                                                   \\
        \hline

        \multirow{9}{1em}{$\Pi$}                           & $\bullet $ Conv2D(filters=50, kernel=5×5, stride=1, padding=2, ReLU)           \\
                                                           & $\bullet $ Batch Normalization                                                 \\
                                                           & $\bullet $ Max Pooling(filters=$2\times2$, stride=2)                           \\
                                                           & $\bullet $ Conv2D(filters=75, kernel=$5\times5$, stride=1, padding=2, ReLU)    \\
                                                           & $\bullet $ Batch Normalization                                                 \\
                                                           & $\bullet $ Max Pooling(filters=$2\times2$, stride=2)                           \\
                                                           & $\bullet $ Conv2D(filters=100, kernel=$3\times3$, stride=1, padding=2, Linear) \\
                                                           & $\bullet $ Batch Normalization                                                 \\
                                                           & $\bullet $ Dense(nb\_neurons=1024, ReLU)                                       \\
        \hline
        \hline

        $\Pi_\tau$                                         & $\bullet $ Dense(nb\_neurons=150, ReLU)                                        \\
        \hline
        \hline

        $\Pi_\sigma$                                       & $\bullet $  Dense(nb\_neurons=150, ReLU)                                       \\
        \hline
        \hline

        \multirow{7}{5em}{$\bar{\Pi}_s$ and $\bar{\Pi}_t$} & $\bullet $ Dense(nb\_neurons=1024, ReLU)                                       \\
                                                           & $\bullet $ Dense(nb\_neurons=6400, ReLU)                                       \\
                                                           & $\bullet $ Conv2D(filters=100, kernel=$3\times3$, stride=1, padding=1, ReLU)   \\
                                                           & $\bullet $ Upsample (scale\_factor=2)                                          \\
                                                           & $\bullet $ Conv2D(filters=50, kernel=$5\times5$, stride=1, padding=2, ReLU)    \\
                                                           & $\bullet $ Upsample(scale\_factor=2)                                           \\
                                                           & $\bullet $ Conv2D (filters=3, kernel=$5\times5$, stride=1, padding=2, Sigmoid) \\
        \hline
        \hline

        \multirow{2}{1em}{$c$}                             & $\bullet $ Dropout(p=0.55)                                                     \\
                                                           & $\bullet $ Dense(nb\_neurons=10, Softmax)                                      \\
        \hline
        \hline

        \multirow{3}{5em}{$r_\tau$ and $r_\sigma$}         & $\bullet $ Gradient Reversal Layer                                             \\
                                                           & $\bullet $ Dense(nb\_neurons=100, ReLU)                                        \\
                                                           & $\bullet $ Dense(nb\_neurons=150, Linear)                                      \\
        \hline
      \end{tabular}
  \end{center}
  \caption{MNIST$\leftrightarrow$USPS networks architectures}
  \label{fig:archi_DA2}
\end{table}

\begin{table}[h]
\small
  \begin{center}
      \begin{tabular}{||c | l||}
        \hline
        Network                                            & Architecture                                                                   \\
        \hline

        \multirow{12}{1em}{$\Pi$}                          & $\bullet $ Conv2D(filters=32, kernel=$5\times5$, stride=1, padding=2, ReLU)    \\
                                                           & $\bullet $ Instance Normalization                                              \\
                                                           & $\bullet $ Max Pooling(filters=$2\times2$, stride=2)                           \\
                                                           & $\bullet $ Conv2D(filters=32, kernel=$5\times5$, stride=1, padding=2, ReLU)    \\
                                                           & $\bullet $ Instance Normalization                                              \\
                                                           & $\bullet $ Max Pooling(filters=$2\times2$, stride=2)                           \\
                                                           & $\bullet $ Conv2D(filters=32, kernel=$3\times3$, stride=1, padding=2, Linear)  \\
                                                           & $\bullet $ Instance Normalization                                              \\
                                                           & $\bullet $ Max Pooling(filters=$2\times2$, stride=2)                           \\
                                                           & $\bullet $ Conv2D(filters=32, kernel=$3\times3$, stride=1, padding=2, Linear)  \\
                                                           & $\bullet $ Instance Normalization                                              \\
                                                           & $\bullet $ Dense(nb\_neurons=1024, ReLU)                                       \\
        \hline
        \hline

        $\Pi_\tau$                                         & $\bullet $ Dense(nb\_neurons=150, ReLU)                                        \\
        \hline
        \hline

        $\Pi_\sigma$                                       & $\bullet $ Dense(nb\_neurons=150, ReLU)                                        \\
        \hline
        \hline

        \multirow{9}{5em}{$\bar{\Pi}_s$ and $\bar{\Pi}_t$} & $\bullet $ Dense(nb\_neurons=1024, ReLU)                                       \\
                                                           & $\bullet $ Dense(nb\_neurons=1024, ReLU)                                       \\
                                                           & $\bullet $ Conv2D(filters=32, kernel=$3\times3$, stride=1, padding=1, ReLU)    \\
                                                           & $\bullet $ Upsample (scale\_factor=2)                                          \\
                                                           & $\bullet $ Conv2D(filters=32, kernel=$3\times3$, stride=1, padding=1, ReLU)    \\
                                                           & $\bullet $ Upsample (scale\_factor=2)                                          \\
                                                           & $\bullet $ Conv2D(filters=32, kernel=$5\times5$, stride=1, padding=2, ReLU)    \\
                                                           & $\bullet $ Upsample(scale\_factor=2)                                           \\
                                                           & $\bullet $ Conv2D (filters=3, kernel=$5\times5$, stride=1, padding=2, Sigmoid) \\
        \hline
        \hline

        \multirow{2}{1em}{$c$}                             & $\bullet $ Dropout(p=0.55)                                                     \\
                                                           & $\bullet $ Dense(nb\_neurons=43, Softmax)                                      \\
        \hline
        \hline

        \multirow{3}{5em}{$r_\tau$ and $r_\sigma$}         & $\bullet $ Gradient Reversal Layer                                             \\
                                                           & $\bullet $ Dense(nb\_neurons=100, ReLU)                                        \\
                                                           & $\bullet $ Dense(nb\_neurons=150, Linear)                                      \\
        \hline
      \end{tabular}
  \end{center}
  \caption{Syn-Signs$\rightarrow$GTSRB networks architectures}
  \label{fig:archi_DA3}
\end{table}

\subsection{Computing resources and code release}
All the experiments from section \ref{sec:xp} were run on a Google Cloud Platform n1-standard-8 virtual machine (8 virtual cores, 30Go RAM, Nvidia P100 GPU).
The code corresponding to the experiments, a list of dependencies, and pre-trained models are available at \url{https://github.com/SuReLI/DiCyR_code}. 
Details about each experiment are reported in Table \ref{table:experiments_details}.

\begin{table}[t]
\begin{center}
\begin{threeparttable}[t]
 \begin{tabular}{||l | l| l | l ||} 
 \hline
 Experiment & Batch size & Epochs & Repetitions\\
 \hline
 \hline
 5.1 SVHN & 64 & 50 (35s/epoch)& 50 \\  
 \hline
 5.1 3D Shapes  & 64 & 50 (15s/epoch) & 5 \\  
 \hline
 5.2 MNIST$\rightarrow$USPS  & 128 & 150 (11s/epoch) & 20 \\  
 \hline
 \ref{sec:xp-usup} USPS$\rightarrow$MNIST & 128 & 150 (11s/epoch) & 20 \\  
 \hline
 \ref{sec:xp-usup} MNIST$\rightarrow$SVHN  & 128 & 50 (40s/epoch) & 50 \\  
 \hline
 \ref{sec:xp-usup} SVHN$\rightarrow$MNIST & 64 & 50 (45s/epoch) & 50 \\  
 \hline
 \ref{sec:xp-usup} Syn-Signs$\rightarrow$GTSRB  & 64 & 150 (65s/epoch) & 10 \\  
 \hline
\end{tabular}
\caption{Experimental setup}
 \label{table:experiments_details}
 \end{threeparttable}
\end{center}
\end{table}

\end{document}